%% file: main.tex
\documentclass[10pt,twocolumn,letterpaper]{article}
\usepackage[pagenumbers]{cvpr}
\input{preamble}
\definecolor{cvprblue}{rgb}{0.21,0.49,0.74}
\usepackage[breaklinks,colorlinks,allcolors=cvprblue]{hyperref}
\title{Why Flat Drafts Stay Flat:\\Region-Mean Latents Govern Texture in Flow-Based Image Generation}
\author{
Zihan Gu$^{1,2}$, Jiangcan Jia$^{3}$, Hua Zhang$^{1,2}$, Lulu Zhao$^{4}$\\
\small$^{1}$Institute of Information Engineering, Chinese Academy of Sciences \;
\small$^{2}$School of Cyber Security, University of Chinese Academy of Sciences \\\small$^{3}$East China Normal University \;
\small$^4$Alibaba Group \;
\small\texttt{\{guzihan\}@iie.ac.cn}
}
\begin{document}
\maketitle
\raggedbottom
\input{section/abstract}
\input{section/teaser}
\input{section/introduction}
\input{section/related_work}
\input{section/mechanism}
\begin{figure*}[!t]
 \centering\includegraphics{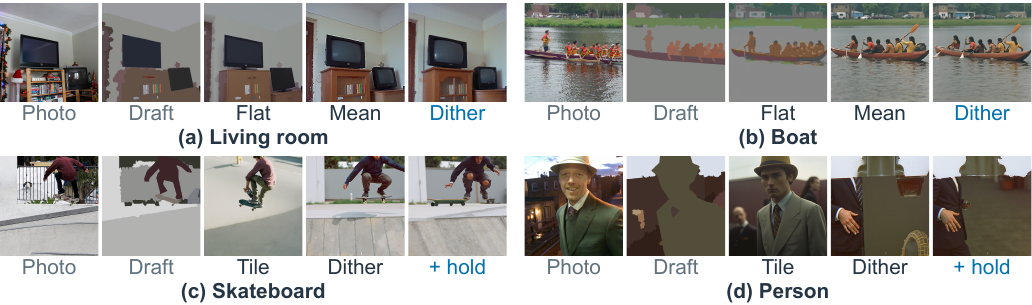}
 \caption{Training-free interventions with fixed prompt and noise within each comparison. (a) Living room: with FlowEdit, mean injection releases fine variation but retains outlined surfaces; dither gives a more photographic rendering. (b) Boat: with RF-Inversion, both mean injection and dither recover boat and water detail. (c) Skateboard: draft corrections retain the gray layout better than Tile. (d) Person: Tile recovers a recognizable subject, while draft corrections misinterpret the silhouette. ``Mean'' injects only the mean field; ``+ hold'' adds boundary holding to dither at scale 1.15. Tile uses strength 0.4. All panels retain the full frames.}
 \label{fig:training-free}
 \vspace{-3pt}
\end{figure*}
\input{section/experiments}
\input{section/conclusion}
\label{main:prose-end}\label{main:end}
\FloatBarrier
\clearpage
\input{section/llm_usage}
{\small\bibliographystyle{ieeenat_fullname}\bibliography{main}}
\clearpage
\appendix
\input{appendxi/construction}
\input{appendxi/metrics}
\input{appendxi/implementation}
\input{appendxi/additional_results}
\FloatBarrier
\onecolumn\twocolumn
\raggedbottom
\allowdisplaybreaks[1]
\begingroup
\setlength{\abovedisplayskip}{7pt plus 2pt minus 2pt}
\setlength{\belowdisplayskip}{7pt plus 2pt minus 2pt}
\input{appendxi/theory}
\endgroup
\end{document}

%% file: preamble.tex
\usepackage[T1]{fontenc}
\usepackage{amsmath,amssymb,amsthm}
\usepackage{graphicx,booktabs,multirow,array}
\usepackage{colortbl,tabularx}
\usepackage{placeins}
\usepackage{stfloats}
\usepackage{needspace}
\usepackage{capt-of}
\usepackage{microtype}
\usepackage{enumitem}
\usepackage{mathtools}
\makeatletter
\newcommand{\tablepanel}[2][\linewidth]{%
  \begin{minipage}[t]{#1}%
  \renewenvironment{table}[1][]{}{}%
  \renewenvironment{table*}[1][]{}{}%
  \def\@captype{table}%
  #2%
  \end{minipage}%
}
\makeatother
\newtheorem{proposition}{Proposition}
\DeclareMathOperator{\Var}{Var}
\DeclareMathOperator{\Cov}{Cov}
\DeclareMathOperator{\tr}{tr}
\DeclareMathOperator{\diag}{diag}
\DeclareMathOperator{\range}{range}
\newcommand{\E}{\mathbb E}
\newcommand{\R}{\mathbb R}
\newcommand{\norm}[1]{\lVert#1\rVert}
\newcommand{\ip}[2]{\langle#1,#2\rangle}
\newcommand{\enc}{\mathcal E}
\newcommand{\dec}{\mathcal D}
\newcommand{\Sone}{S_1}
\newcommand{\Stwo}{S_2}
\newcommand{\Tgeom}{T_{\mathrm{geom}}}
\newcommand{\Tpool}{T_{\mathrm{pool}}}
\definecolor{paperblue}{HTML}{0072B2}
\definecolor{paperorange}{HTML}{D55E00}
\definecolor{paperink}{HTML}{253544}
\definecolor{papergray}{HTML}{61717D}
\definecolor{paperrule}{HTML}{000000}
\definecolor{paperwash}{HTML}{EEF5FA}
\definecolor{paperlight}{HTML}{F5F7F9}
\newcommand{\tblhead}[1]{\textbf{#1}}
\newcommand{\tblgroup}[2]{\multicolumn{#1}{l}{\color{black}\textit{#2}}\\}
\newcommand{\tblbest}[1]{\textbf{#1}}
\newcommand{\tblsecond}[1]{\underline{#1}}
\newcommand{\tblfocus}[1]{\textbf{#1}}
\newcommand{\tblnote}[1]{#1}
\setlist[itemize]{leftmargin=*,topsep=3pt,itemsep=2pt,parsep=0pt}

\newcommand{\papertablestyle}{%
\renewcommand{\arraystretch}{1.06}
\setlength{\heavyrulewidth}{0.75pt}
\setlength{\lightrulewidth}{0.34pt}
\setlength{\aboverulesep}{1.8pt}
\setlength{\belowrulesep}{2.2pt}}

%% file: section/abstract.tex
\begin{abstract}
Flat color-block drafts specify layout and regional color while leaving interior detail unspecified, yet pretrained flow-based generators can preserve their flat appearance even under photorealistic prompts. We trace this behavior to a cross-scale mechanism linking fine pixel variation to regional means in variational autoencoder (VAE) latent space. Flattening photo regions systematically shifts their latent means, with offsets concentrated along a shared channel direction within each backbone. Controlled field exchanges in FLUX.1-dev and Qwen-Image establish the effect of this displacement on texture generation: adding the photo-derived regional mean field to a flat latent restores texture, whereas adding its complementary field alone largely preserves flatness. A second-order encoding expansion explains how encoder curvature can convert zero-mean perturbations into mean shifts. An analytically solvable conditional Gaussian flow shows how regional noise averaging enables near-unit endpoint sensitivity at high signal-to-noise ratios. This mechanism motivates training-free rendering through pixel dither before encoding, mean-field scaling, and early boundary holding. Paired low-rank adaptation (LoRA) with draft conditioning improves content recovery and produces photo-like regional means under the standard velocity objective. Explicit regional mean guidance remains effective after adaptation, continuously controlling texture and improving color fidelity. These findings connect encoding shifts to persistent generation behavior and support interventions evaluated for texture, spatial fidelity, and semantic recovery.
\end{abstract}

%% file: section/teaser.tex
\begin{figure}[!t]
 \centering\includegraphics{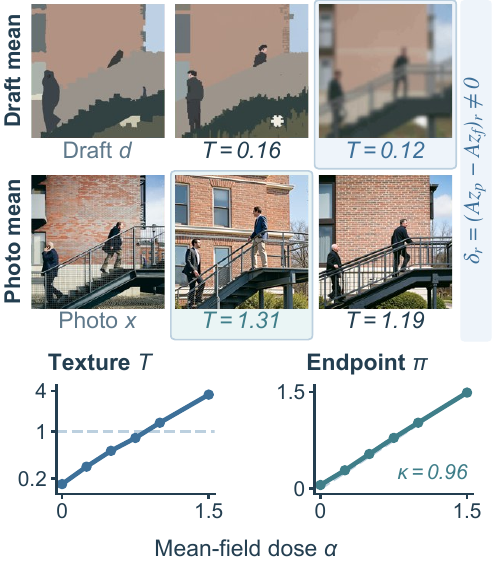}
 \caption{\textbf{Flatness follows regional latent means.} FLUX.1-dev with matched prompt/noise. Left: draft $d$ and photo $x$, with equal regional RGB means before quantization. Their VAE latents $z_f,z_p$ have different regional means, measured by $A$. In the output grid, rows select draft/photo mean fields and columns select draft/photo complementary fields ($b=0/1$), using the soft-mask operator $P$. The photo complement alone leaves surfaces flat; the photo mean field restores texture. Matrix $T$: geometric texture ratios over 300 images. Curves: 100 images; dashed lines mark source texture $T=1$ and identity $\pi=\alpha$.}
 \label{fig:intro}
\end{figure}

%% file: section/introduction.tex
\section{Introduction}\label{sec:introduction}\label{sec:intro}
A flat color-block draft provides a compact visual specification: region boundaries indicate object extent, regional colors constrain appearance, and uniform interiors leave shading and texture open for synthesis. Image-to-image generation and diffusion-based dequantization offer ways to complete such simplified inputs~\cite{meng2022sdedit,vavilala2025dequantization}. SDEdit adds noise to a reference before denoising it with a generative prior. Applied to the latent flow generators studied here, this strategy exposes a persistent tension. Moderate noise preserves smooth interiors even when the target prompt requests photorealistic detail; stronger noise permits richer texture but weakens regional color and boundary fidelity. \Cref{fig:intro} previews the latent-mean mechanism behind this behavior.

Latent image generators combine learned image representations~\cite{rombach2022ldm} with generative dynamics, including flow matching and rectified flow~\cite{lipman2022flowmatching,liu2022rectifiedflow}. In the models studied here, a variational autoencoder (VAE) supplies the image representation. At the pixel level, replacing each photo region by its average RGB color preserves its mean, apart from quantization, while removing interior variation. The nonlinear encoder maps these pixel changes into both regional latent averages and within-region structure. We find that the resulting regional mean displacement persists along the generation trajectory and constrains the detail synthesized from the draft.

Fine pixel variation is encoded across scales into \emph{regional VAE latent means}. Replacing photo regions with flat colors systematically displaces their average latent vectors. Within each backbone, the photo-minus-draft offsets concentrate along a shared channel direction, which accounts for $78\%$ of held-out offset energy in FLUX.1-dev~\cite{bflFlux} and $95\%$ in Qwen-Image~\cite{qwen2025image}. Synthetic color-block perturbations support a nonlinear encoding explanation. For zero-mean perturbations, a local second-order expansion cancels the linear term in expectation, while encoder curvature can convert pixel variance into a coherent latent mean shift. The displacement therefore appears before sampling begins.

The key test separates regional statistics from spatial detail. We independently exchange the regional mean fields and complementary fields of paired draft and photo latents, keeping prompts and sampling noise fixed. In both backbones, the photo complementary field leaves surfaces smooth when combined with the draft mean field, whereas the photo mean field releases texture with the draft complement. Equal-norm random fields yield texture gains near zero or below, and object-only interventions concentrate texture changes on treated regions. The regional mean field therefore controls whether the trajectory synthesizes detail, beyond the fine spatial information already present in its starting latent.

The injected mean coordinate also survives sampling. Across graded interventions, generated texture rises with the supplied mean field, and terminal photo-relative mean position tracks the nominal field dose with slopes of $0.96$ in FLUX and $0.90$ in Qwen. An analytically solvable conditional Gaussian flow explains how strong retention can arise: spatial averaging suppresses independent sampling noise according to effective region size, and endpoint sensitivity approaches one at high signal-to-noise ratio (SNR). A complementary two-mode Gaussian model shows how a reference mean shift can outweigh text-conditioned prior odds for textured versus flat appearance. Together, the encoding and trajectory analyses explain how an input intended to leave detail open can impose a persistent smooth-appearance constraint.

The mechanism suggests a training-free correction computed from the draft at inference. Adding low-amplitude pixel dither before VAE encoding induces a useful regional mean shift and enables detail generation with SDEdit-style initialization, FlowEdit~\cite{kulikov2025flowedit}, and RF-Inversion~\cite{rout2024rfinversion}. Field ablations attribute the main texture gain to the dither-induced mean field; transfer experiments also show that the complementary field can matter for photographic appearance. To preserve the draft's spatial specification, we combine mean-field scaling with early boundary holding, inspired by known-region enforcement during diffusion inpainting~\cite{lugmayr2022repaint}. The resulting recipe improves regional color and edge fidelity over the tested Tile ControlNet baseline~\cite{zhang2023controlnet,shakkerControlNetUnionPro} on 300 matched confirmation images (\cref{tab:main-rendering}).

Texture restoration and semantic recovery remain distinct objectives. Flat silhouettes can be ambiguous even when their surfaces acquire detail. Paired training with low-rank adaptation (LoRA)~\cite{hu2021lora} and draft conditioning improves object recovery. An orthogonal decomposition of the standard velocity loss exposes its regional-mean and within-region terms; prediction and condition probes show that the adapted predictor reduces regional mean errors and responds to moderate condition shifts near the photo-level target (\cref{fig:learning}). The same coordinate remains controllable after adaptation: explicit mean guidance varies texture with the adapter fixed (\cref{fig:lora-dial}) and lowers regional color error from $7.1$ to $4.0$. Adaptation improves content recovery; inference-time guidance controls appearance and regional color.

Our contributions are three connected claims:
\begin{itemize}
 \item \textbf{Mechanism of persistent flatness.} We identify a shared regional latent displacement induced by flattening and establish its effect on texture through field exchanges, direction and locality controls, and dose-response measurements in two flow-based backbones.
 \item \textbf{Regional operators and analytical models.} We distinguish soft-mask field interventions from exact orthogonal projections, and derive models of nonlinear encoding shifts and high-SNR retention along conditional Gaussian flows.
 \item \textbf{Mechanism-guided rendering and adaptation.} We use dither and boundary holding for training-free correction, and show that regional mean correction and explicit texture control remain relevant after paired LoRA adaptation, with appearance, layout, and content evaluated separately.
\end{itemize}

%% file: section/related_work.tex
\section{Related Work}\label{sec:related}
Draft rendering combines appearance synthesis with preservation of an incomplete visual specification. SDEdit~\cite{meng2022sdedit} adds noise to a reference before denoising, and ILVR~\cite{choi2021ilvr} imposes coarse image constraints during sampling. Dequantization and color transfer~\cite{vavilala2025dequantization} reconstruct natural images from quantized colors; the Color Diffusion generator used by Mao et al.~\cite{mao2025colorillusions} learns to render colored sketches. We analyze the regional latent statistic that changes when a detailed reference is replaced by a flat draft.

Spatial conditioning provides several interfaces for retaining reference structure. ControlNet~\cite{zhang2023controlnet} and T2I-Adapter~\cite{mou2023t2iadapter} learn auxiliary conditions, whereas Plug-and-Play Diffusion Features~\cite{tumanyan2022pnp} and FreeControl~\cite{mo2024freecontrol} reuse pretrained representations during generation. FlowEdit~\cite{kulikov2025flowedit} transports between source and target prompts without inversion; RF-Inversion~\cite{rout2024rfinversion} formulates semantic inversion and editing through rectified stochastic dynamics. Kontext~\cite{bfl2025kontext} supports image-context editing, and Redux~\cite{bflFluxRedux} supplies image-variation conditioning. RePaint~\cite{lugmayr2022repaint} enforces known image regions during denoising. We compare these interfaces for draft rendering and use boundary holding to preserve layout during latent correction.

Latent statistics and frequency structure influence generative behavior. Spectrum Matching~\cite{ning2026spectrummatching} studies the diffusability of learned representations, and DeBaT~\cite{medi2025debat} separates frequency bands in image tokenization. FreeInit~\cite{wu2024freeinit} and noise-schedule analysis~\cite{lin2023noiseschedules} connect generation failures to initialization and sampling statistics. FlexiEdit~\cite{koo2024flexiedit} refines latent frequencies for editing, and FreeU~\cite{si2023freeu} changes the balance of intermediate diffusion features. PixSDS~\cite{skorokhodov2026pixsds} examines pixel directions weakly constrained by a latent objective. Our analysis holds the encoder and generator fixed and traces a fine-scale pixel change into a regional latent average, then into generated texture.

VAE-LFA~\cite{wang2026vaelfa} is a close statistical intervention: it aligns low-frequency latent statistics to reduce drift in multi-turn editing. The statistic's desired behavior depends on the reference. Preserving an edited image's statistics can maintain its appearance; preserving a flat draft's statistics can sustain unwanted flatness. Here, a single-trajectory intervention deliberately changes regional means, with its direction and dose tested against controls. The image partition defines our averaging operator, so each measured mean follows the spatial support of its region.

Intervention-based interpretation links representations to generative effects. Asyrp~\cite{kwon2023asyrp}, Riemannian latent-space analysis~\cite{park2023riemann}, and sparse-autoencoder studies~\cite{tinaz2025emergence} analyze interpretable directions or concepts in diffusion models. Concept Sliders~\cite{gandikota2023sliders} learns low-rank parameter directions for continuous control. At the training level, DiT~\cite{peebles2023dit} establishes scalable latent transformers, while REPA~\cite{yu2025repa} aligns their representations with visual features. We intervene on a directly measurable VAE-region statistic. This coordinate connects a frozen-model diagnosis, a training-free correction, and the behavior of a learned draft renderer at the level of regional latent statistics.

%% file: section/mechanism.tex
\section{Mechanism and Guided Rendering}\label{sec:mechanism}\label{sec:rendering}
Regional latent means carry smooth-appearance information from a flat draft. We trace its nonlinear encoding and trajectory retention, then use the same operators for draft-only correction and learned control. \Cref{fig:method} summarizes the mechanism and the resulting controls.
\input{section/method_figure}

\subsection{Regional subspace and nonlinear encoding shift}\label{sec:representation}
For an image $x$ partitioned into $R$ fixed regions, pixel averaging $A_{\rm px}$ and filling $L_{\rm px}$ define $d^\circ=L_{\rm px}A_{\rm px}x$, with $A_{\rm px}(x-d^\circ)=0$. Let $d$ be its eight-bit version (\cref{app:construction}). Deterministic VAE encoding $\enc$ and decoding $\dec$ use normalized latent units: $z_f=\enc(d)$ and $z_p=\enc(x)\in\R^{N\times C}$, for $N$ cells and $C$ channels.

Pooled masks give memberships $M\in\R^{R\times N}$, with $M_{ri}\geq0$ and $\sum_rM_{ri}=1$. After removing empty regions, let $a_r=\sum_iM_{ri}$ and $W=\diag(a_r)$. Define the regional operators by
\begin{equation}\label{eq:operators}
\begin{alignedat}{2}
 A&=W^{-1}M, &\qquad S&=M^\top,\\
 P&=SA,      &       B&=AS.
\end{alignedat}
\end{equation}
$A$ pools regional means and $S$ spreads regional vectors. We call $P\Delta z$ the mean field and $(I_N-P)\Delta z$ the complementary field. The exact spatial projectors $\Pi=A^\dagger A$ and $Q=I_N-\Pi$ use the Moore--Penrose inverse and satisfy $AQ=QS=0$ (\cref{app:operators}); the soft-mask operator generally has $P^2\neq P$.

For the photo offset $\Delta z_p=z_p-z_f$, let $\delta=A\Delta z_p$. Two regional coordinates measure its shared-direction and photo-relative components:
\begin{equation}\label{eq:offset}
\begin{aligned}
 \rho_r(z)&=\ip{(Az-Az_f)_r}{u},\\
 \pi_r(z)&=\frac{\ip{(Az-Az_f)_r}{\delta_r}}{\norm{\delta_r}^2}.
\end{aligned}
\end{equation}
Each backbone fits a unit $u\in\R^C$ by normalizing the average photo-minus-draft offset on a separate set. For $\delta_r\neq0$, $\pi_r(z_f)=0$ and $\pi_r(z_p)=1$. The channel components $\delta_r^\parallel=\rho_r(z_p)u$ and $\delta_r^\perp=(I_C-uu^\top)\delta_r$ sum to $\delta_r$; Q1 tests their texture and color effects.

For a fixed mask/channel, put $f(y)=[A\enc(y)]_{rj}$. Let $\xi$ have zero mean, covariance $\Sigma$, and finite third moment. A uniform bound on third derivatives along perturbation segments gives, for $\eta\geq0$,
\begin{equation}\label{eq:curvature}
\begin{aligned}
 &\E\big[f(d+\eta\xi)-f(d)\big]\\
 &\quad=\frac{\eta^2}{2}\tr\!\big(H_f(d)\Sigma\big)
       +O\!\big(\eta^3\E\norm{\xi}^3\big).
\end{aligned}
\end{equation}
The linear term cancels; curvature can convert fine variation into a mean shift. Q1 tests alignment with $u$; \cref{app:encoding} gives the remainder bound and a Fourier example.

\subsection{Trajectory interventions and analytical retention}\label{sec:interventions}\label{sec:background}
Fix text $c$, noise $\varepsilon\sim\mathcal N(0,I)$, masks, model, and schedule. From noise level $s$, sampler $\Phi^c$ gives the endpoint and decoded output
\begin{equation}\label{eq:render}
\begin{aligned}
 z_{\rm out}(z)&=\Phi^c_{s\to0}\!\big((1-s)z+s\varepsilon\big),\\
 Y(z)&=\dec\big(z_{\rm out}(z)\big).
\end{aligned}
\end{equation}
For any reference offset $\Delta z$, the common intervention is
\begin{equation}\label{eq:factorial}
 z(a,b;\Delta z)=z_f+\big[aP+b(I_N-P)\big]\Delta z.
\end{equation}
Writing $\delta_\Delta=A\Delta z$, its measured mean and exact spatial residual obey
\begin{equation}\label{eq:softmeans}
\begin{aligned}
 Az(a,b;\Delta z)&=Az_f+\big[aB+b(I_R-B)\big]\delta_\Delta,\\
 Qz(a,b;\Delta z)&=Qz_f+bQ\Delta z.
\end{aligned}
\end{equation}
Mean-field changes preserve the exact spatial residual $Qz$; soft complements also affect means through $(I_R-B)\delta_\Delta$. For hard, disjoint latent masks, $B=I_R$ and the two fields exchange regional means and spatial residuals independently. With pooled masks, \cref{eq:softmeans} specifies the coupling for every intervention. Q1 uses $\Delta z_p$, with $a,b\in\{0,1\}$ for the factorial $z_{ab}$ and $(a,b)=(\alpha,0)$ for dose response. Q2 uses the dither offset.

Let $T(Y)$ measure source-relative within-region texture and $\ell_{ab}=\log\max\{T(Y(z_{ab})),\epsilon_T\}$. The matched main effects are
\begin{equation}\label{eq:effects}
\begin{aligned}
 e_M&=\frac12[(\ell_{10}-\ell_{00})+(\ell_{11}-\ell_{01})],\\
 e_C&=\frac12[(\ell_{01}-\ell_{00})+(\ell_{11}-\ell_{10})].
\end{aligned}
\end{equation}
Bars denote image averages; \cref{app:texture} specifies $\epsilon_T$ and texture measurement. Retention $\kappa$ is the fitted slope of endpoint $\pi_r$ versus nominal dose $\alpha$.

Regional averaging reduces independent unit Gaussian latent noise, in each unit channel direction, to variance
\begin{equation}\label{eq:noise}
 \nu_r=\frac{\sum_i M_{ri}^2}{a_r^2}=\frac1{n_{{\rm eff},r}}.
\end{equation}
An ideal flow learns a conditional-mean velocity~\cite{lipman2022flowmatching,liu2022rectifiedflow}; its clean-prediction identity is derived in \cref{app:posterior}.

Consider an autonomous Gaussian coordinate with conditional variance $\tau^2>0$ and independent Gaussian noise variance $\nu_r$, under the linear noising path of \cref{eq:render}. For $0<s<1$, its ideal marginal-flow endpoint $q_{\rm out}$ has fixed-noise sensitivity to reference $q_{\rm ref}$
\begin{equation}\label{eq:gaussian-retention}
\begin{aligned}
 \kappa_{\rm flow}:=\frac{\partial q_{\rm out}}{\partial q_{\rm ref}}
 &=\sqrt{\frac{\mathrm{SNR}}{1+\mathrm{SNR}}},\\
 \mathrm{SNR}&=\frac{(1-s)^2\tau^2}{s^2\nu_r}.
\end{aligned}
\end{equation}
This equals $\sqrt{\kappa_{\rm post}}>\kappa_{\rm post}$ for single-prediction sensitivity $\kappa_{\rm post}=\mathrm{SNR}/(1+\mathrm{SNR})$; both approach one at high SNR (\cref{app:retention-model}). Effective region size fixes the averaged noise variance; conditional variance and coordinate coupling also govern retention. Q1 tests area dependence.

For Gaussian textured/flat modes with mean difference $\Delta>0$ and common noised variance $V_s$, a reference shift $h$ changes posterior log odds by $(1-s)^2\Delta h/V_s$ at fixed noise, which can outweigh text-conditioned prior odds (\cref{app:appearance-mixture}).

\subsection{Training-free correction: dither and boundary holding}\label{sec:dither}
Dither obtains a displacement from the draft alone. Add independent Gaussian pixel noise of standard deviation $\eta$, clip, and quantize to form $\widetilde d_\eta$. Set $z_d=\enc(\widetilde d_\eta)$, $\Delta z_d=z_d-z_f$, and $\delta_d=A\Delta z_d$. Binary field exchanges in \cref{eq:factorial} now isolate the two dither fields. Keeping its complement and scaling its mean field gives $z(\alpha,1;\Delta z_d)$.

Boundary holding constrains layout during early sampling. Let $\mathcal B$ contain packed image tokens intersecting a region boundary, and let $\sigma_i$ be the decreasing noise schedule. With $z_{\rm init}(\alpha)=z(\alpha,1;\Delta z_d)$ and the same noise $\varepsilon$, define a noised reference path and replace selected boundary tokens after each Euler update:
\begin{equation}\label{eq:dither}
\begin{aligned}
 z_{\rm init}(\alpha)&=z_d+(\alpha-1)S\delta_d,\\
 z_i^{\rm ref}&=(1-\sigma_i)z_{\rm init}(\alpha)+\sigma_i\varepsilon,\\
 z_{i+1}\big|_{\mathcal B}&\leftarrow z_{i+1}^{\rm ref}\big|_{\mathcal B}.
\end{aligned}
\end{equation}
Plain dither uses $\alpha=1$ and leaves all tokens free throughout sampling. Interior tokens follow the model throughout; boundary tokens are released after the holding interval. Packing, holding steps, transfer interfaces, and computation are specified in \cref{app:implementation}.

\subsection{Paired adaptation: learning and controlling regional means}\label{sec:learning}
A low-rank adapter $\phi$ learns from draft/photo pairs using extra draft tokens and a frozen backbone $\theta$. For $Z_\sigma=(1-\sigma)z_p+\sigma\varepsilon$, write $v=v_{\theta,\phi}(Z_\sigma,\sigma,c,z_f)$ and $\widehat z_0=Z_\sigma-\sigma v$. The errors $e_v=v-(\varepsilon-z_p)$ and $e_0=\widehat z_0-z_p=-\sigma e_v$ give, for $\sigma>0$, the ordinary loss split in Frobenius norm:
\begin{equation}\label{eq:loss-split}
\begin{aligned}
 \mathcal L(\phi)&=\E\norm{e_v}_F^2\\
 &=\E\!\left[\sigma^{-2}\big(\norm{\Pi e_0}_F^2+\norm{Qe_0}_F^2\big)\right].
\end{aligned}
\end{equation}
Ordinary velocity supervision thus penalizes regional mean errors; \cref{app:operators} gives their exact weighted norm.

Q3 measures $Ae_0$ along $u$ and in $u^\perp$, plus a within-region readout (\cref{app:probe-metrics}). The complementary-field probe uses $I_N-P$ and reports MSE; the regional probes report area-weighted RMS errors. The exact loss identity uses $\Pi$ and $Q$. The condition probe feeds $z_{\rm cond}(\beta)=z_f+\beta S\delta_d$ at fixed noisy state and prompt. Its area-weighted output shift $\rho(\widehat z_0)$ tests a fixed additive correction versus a photo-level target (\cref{app:probe-conventions}).

At inference, mean guidance takes an area-weighted least-squares gradient step (\cref{app:guidance-derivation}):
\begin{equation}\label{eq:guidance}
\begin{aligned}
 m^\star&=Az_f+\alpha\delta_d,\\
 \widehat z_0'&=\widehat z_0+\gamma S(m^\star-A\widehat z_0).
\end{aligned}
\end{equation}
For fixed attainable $m^\star$, isolated steps converge when $0<\gamma<2$, preserving $Q\widehat z_0$. Sampling interleaves these corrections with nonlinear updates through $v'=(z_\sigma-\widehat z_0')/\sigma$. Varying $\alpha$ controls appearance with the adapter and draft fixed.

%% file: section/method_figure.tex
\begin{figure*}[t]
 \centering\includegraphics[width=\textwidth]{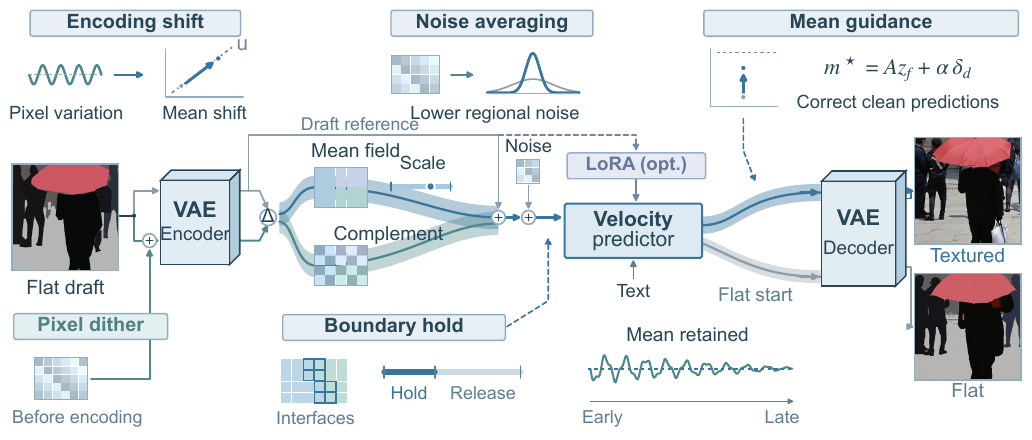}
 \caption{\textbf{Regional means connect encoding, trajectory retention, and rendering control.} Pre-encoding dither produces a latent displacement; its mean field is scaled while its spatial complement is retained. The lanes represent the operational split $P$ and $I-P$. Regional noise averaging motivates high mean retention. Boundary tokens are held early and then released; optional draft-conditioned LoRA and mean guidance act on the same velocity predictor. The guidance target uses $\delta_d=A(z_d-z_f)$. Real FLUX.1-dev panels show one saved case: the draft (left), dither with mean scale $1.15$ and boundary holding (upper right), and the flat-start rendering (lower right). These panels illustrate the training-free branch. Paths, fields, and coordinate insets are schematic; the gray path denotes a separate uncorrected run.}
 \label{fig:method}
 \vspace{-4pt}
\end{figure*}

%% file: section/experiments.tex
\section{Experiments}\label{sec:experiments}
We ask three questions: (Q1) why denoising preserves a flat draft, (Q2) whether the mechanism enables training-free rendering, and (Q3) whether it remains active after paired LoRA adaptation. FLUX.1-dev and Qwen-Image receive mean-color drafts from COCO panoptic regions. The mechanism and confirmation sets contain 300 disjoint images each. Interventions within a method share prompts and sampling noise. Splits and execution settings are in \cref{app:construction,app:implementation}.

We measure source-relative interior texture ($\Tgeom$: geometric aggregation; $\Tpool$: pooled aggregation), regional color error ($\Delta E$), and boundary alignment (Edges). Classes and Objects measure semantic agreement and instance recovery; HPSv3 measures predicted preference. Joint success $\Stwo$ requires target style, color fidelity, semantic agreement, and preference on the same image. Texture describes the amount of fine variation relative to the source; style, fidelity, and content scores assess the resulting rendering. Detailed definitions, thresholds, and estimators are collected in \cref{app:metrics}.

\subsection{Why does denoising preserve a flat draft?}\label{sec:exp-mechanism}
Flattening changes a coherent regional latent statistic, and the generation trajectory retains it strongly enough to constrain detail. Encoding, field exchange, and dose response test this mechanism (\cref{tab:causal-summary}). The photo-field factorial uses 300 FLUX and 100 Qwen images; direction and dose controls use 100 per model, near starting noise 0.82.

\input{tables/causal_summary_main}

The shared direction contains 78\% of held-out photo-minus-draft offset energy in FLUX and 95\% in Qwen. Fine perturbations of synthetic color blocks induce shifts aligned with the natural-photo offset, with cosine similarities 0.93 and 0.95; equal-magnitude uniform color changes produce much smaller shifts. These measurements connect the nonlinear response in \cref{eq:curvature} to the encoders' sensitivity to fine variation (\cref{fig:encoder,tab:mechanism}). The mean displacement appears during encoding, before any generation trajectory.

The field exchange tests whether missing spatial detail explains persistent flatness. Keeping the draft mean field and supplying the photo complement leaves texture ratios at 0.12 in FLUX and 0.11 in Qwen. Replacing only the regional mean field raises them to 1.31 and 0.95. Averaging each factor's effect over both settings of the other gives log-texture effects of 2.22 and 1.95 for the mean field, versus $-0.20$ for the complement in either backbone. The generator synthesizes detail from a mean-field shift with the draft complement fixed. The photo complement paired with the draft mean field preserves the low-texture regime.

Direction and locality controls distinguish this effect from arbitrary latent disturbance. Equal-norm random fields have log-texture gains near zero or below, and object-only interventions concentrate the response on treated objects. The shared-direction component alone yields texture 0.88 in FLUX and 0.84 in Qwen, but color errors of 13.5 and 9.0; the orthogonal component alone yields FLUX texture 0.20 (\cref{tab:controls}). Texture recovery concentrates along the shared direction; color fidelity also depends on the remaining channel components. Directly decoding a shifted flat latent produces little fine texture: the generative trajectory supplies the detail.

\input{tables/dither}
\input{tables/interventions_main}


\input{tables/main_rendering}

\begin{figure*}[!t]
 \centering\includegraphics{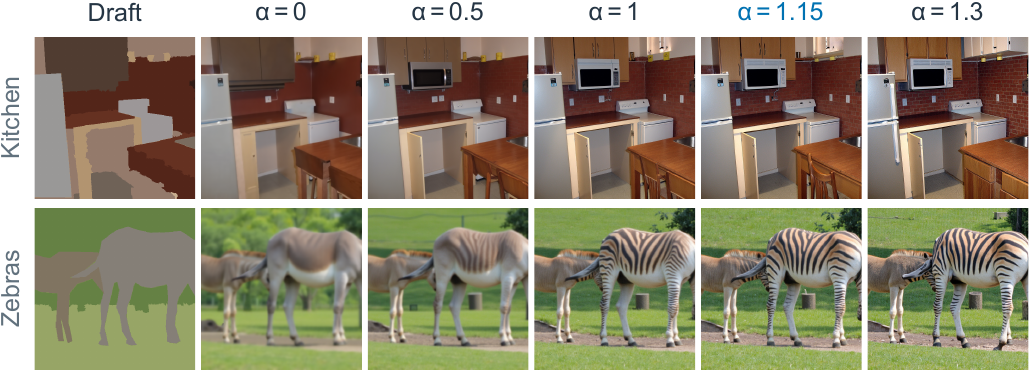}
 \caption{Mean-target control with LoRA, draft, prompt, and noise fixed. All five tested scales are shown. Increasing the target reveals surface detail alongside geometry changes. Blue marks the selected scale 1.15.}
 \label{fig:lora-dial}
 \vspace{-3pt}
\end{figure*}

\begin{figure}[!t]
 \centering\includegraphics{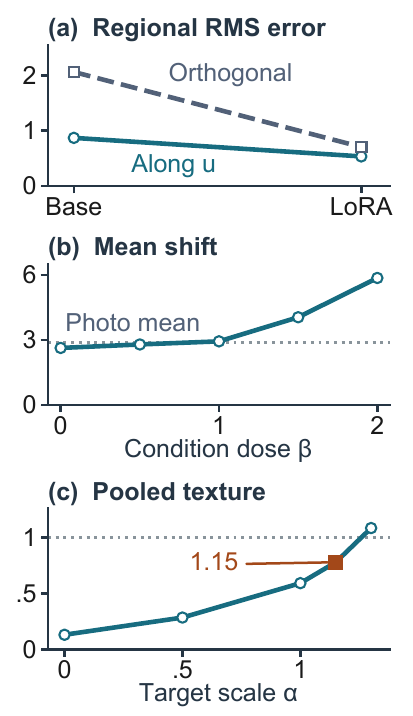}
 \caption{Regional correction and control on 300 images. (a) Regional RMS errors: text-only base and draft-conditioned LoRA. (b) Condition-dose response. Both probes use pure noise and guidance input one. (c) Pooled texture with fixed LoRA and guidance input 3.5; square: selected scale 1.15.}
 \label{fig:learning}
 \vspace{-8pt}
\end{figure}

Graded interventions test whether this regional constraint survives the complete trajectory. Increasing the injected mean dose progressively raises texture, and regressing terminal mean position on dose gives slopes of 0.96 for FLUX and 0.90 for Qwen. These slopes measure sensitivity to the nominal coefficient $\alpha$ in the implemented field intervention. The actual initial regional coordinate follows \cref{eq:softmeans}; terminal measurements establish strong dose dependence after detail synthesis. Higher starting noise weakens retention, whereas the standard-start response is similar across region sizes and area-only fits explain little variation (\cref{tab:retention}). Noise averaging provides an analytical account of strong retention; the measured slopes show that a draft's smooth-appearance constraint persists through sampling.

\subsection{Can the mechanism guide training-free rendering?}\label{sec:exp-free}
A draft-derived mean shift releases detail using the frozen generator and draft-only inference inputs. Q2 substitutes the dither offset $\Delta z_d$ into the same field operator (\cref{tab:dither}), to localize the effect on mechanism images, then tests transfer and rendering quality on the disjoint confirmation set.

The field ablation localizes the texture gain (\cref{tab:dither}). On the same 100 mechanism images, the mean-only start gives $\Tpool=0.72$, versus 0.12 for the complement-only start and 0.17 for the flat start. Full dither is shown separately over all 300 mechanism images. Matched controls then separate boundary holding from scale: at $\alpha=1$, holding raises Edges from 0.18 to 0.33 and lowers color error from 4.3 to 3.2, while reducing $\Tgeom$ from 0.68 to 0.58. Raising $\alpha$ to 1.15 with holding fixed recovers texture to 0.82 (\cref{tab:boundary-ablation}). Holding improves spatial fidelity; scaling restores detail at a measurable fidelity cost.

The intervention transfers to FlowEdit and RF-Inversion, with different roles for the complementary field (\cref{tab:interventions}; \cref{fig:training-free}(a,b)). On 100 matched images, RF-Inversion reaches $\Stwo=0.34$ with means alone and 0.30 with full dither. FlowEdit reveals a sharper distinction: the mean-only intervention raises measured texture, yet full dither increases photo-style rate from 0.26 to 0.79 and joint success from 0.05 to 0.32. The paired success gain is 0.27 (\cref{tab:paired}). Full dither has slightly lower texture magnitude, so its advantage reflects photographic rendering beyond the high-frequency variation restored by the mean field.

Input substitution has different effects in the other interfaces. Kontext already produces substantial detail but changes regional colors and layout; Redux acquires a larger projected shift with little texture gain. Joint success stays at its baseline level or decreases under the tested dither substitution (\cref{tab:interface-quality}). These outcomes distinguish releasing a texture constraint from preserving the information specified by the draft.

Dither also raises Qwen texture from 0.18 to 0.55 and photo-target joint success from 0.02 to 0.15 on 300 images, with worse spatial fidelity (\cref{tab:qwen}). Anime results and all ControlNet modes are in \cref{tab:full-anime,tab:controlnet-anime,tab:controlnet-matched}. These comparisons expose the content ambiguity that motivates paired adaptation.

On 300 images, scale-and-hold improves color and boundary fidelity over plain dither, with lower preference and similar joint success (\cref{tab:main-rendering}; \cref{fig:training-free}(c,d)). Against the tuning-selected Tile ControlNet, color error falls from 12.4 to 3.3 and Edges rises from 0.22 to 0.33; Tile recovers more objects and scores higher on HPSv3. Paired contrasts support both sides of this tradeoff (\cref{tab:paired}).

\vspace{-6pt}
\subsection{Does a draft LoRA remain consistent with the mechanism?}\label{sec:exp-lora}

The adapted predictor corrects regional means, and a fixed adapter remains sensitive to the same appearance coordinate. The FLUX LoRA is trained on 2,221 draft/photo pairs for 1,500 steps with the ordinary velocity loss. All final-checkpoint probes, five-dose controls, and quality comparisons below use the same 300 confirmation images.

The prediction probe connects adaptation to the regional term in the loss decomposition (\cref{eq:loss-split}). From pure noise, the base predictor's along-$u$ RMS error is 0.87, compared with 0.53 for the adapted predictor with draft tokens; orthogonal regional error is 2.06 versus 0.71. The implemented within-region MSE changes less, from 0.337 to 0.310 (\cref{fig:learning}(a), \cref{tab:learning}). The resulting predictor therefore makes a substantial regional correction alongside its content changes.

The condition probe tests how that correction responds to shifted draft tokens. As $\beta$ increases from zero to one, predicted shift moves only from 2.63 to 2.93, near the source-photo level 2.89; at larger doses it rises more strongly (\cref{fig:learning}(b)). This compressed low-dose response places the predictions near the photo-level target over that range. The full condition probe and the original training-checkpoint trajectory are in \cref{tab:condition-probe,fig:learning-history}.

A direct intervention on the fixed adapter establishes continued controllability. Targeting flat means suppresses kitchen surfaces and zebra markings; increasing only the target restores detail (\cref{fig:lora-dial}). Across 300 images, $\Tpool$ rises through 0.13, 0.29, 0.59, 0.78, and 1.09 over the five scales (\cref{fig:learning}(c)). An initialization control agrees: the same adapter gives texture 0.08 from a flat image-to-image start and 0.35 from a dithered start (\cref{tab:lora-readouts}). The regional appearance constraint remains active after learned content recovery.

The matched quality comparison separates adaptation from inference-time guidance (\cref{tab:main-rendering}). Relative to the training-free method, both LoRA variants recover more objects. With the adapter fixed, guidance reduces color error by 3.15 and raises HPSv3 by 1.63; joint success increases from 0.38 to 0.48, with a paired gain of 0.10. Object recovery is slightly lower and boundary alignment also decreases slightly. Thus the learned renderer supplies stronger content recovery, while explicit mean guidance remains a useful control over appearance and regional color.

%% file: tables/causal_summary_main.tex
\begin{table}[tp]
\centering\normalsize
\setlength{\tabcolsep}{4pt}
\papertablestyle
\arrayrulecolor{black}
\caption{Photo-field effects and endpoint retention. Factorial: 300 FLUX and 100 Qwen images; dose regression: 100 per model. Effects are on log texture; $\kappa$ is the terminal-position slope. \tblnote{Bold emphasizes the mean-field effect and retention.}}
\label{tab:causal-summary}
\begin{tabular}{@{}lrrr@{}}
\toprule
\tblhead{Model} & \tblhead{$\bar e_M$} & \tblhead{$\bar e_C$} & \tblhead{$\kappa$} \\
\midrule
FLUX & \tblfocus{2.22} & -0.20 & \tblfocus{0.96} \\
Qwen & \tblfocus{1.95} & -0.20 & \tblfocus{0.90} \\
\bottomrule
\end{tabular}
\end{table}

%% file: tables/dither.tex
\begin{table}[t]
\centering
\normalsize
\setlength{\tabcolsep}{3pt}
\papertablestyle
\arrayrulecolor{black}
\caption{Dither-field ablation in FLUX, photo target. Controls share 100 mechanism images; full dither uses 300. $\Tpool$: pooled texture; $\rho$: re-encoded output shift. Soft-mask fields. \tblnote{Bold: mean-only control.}}
\label{tab:dither}
\begin{tabular}{@{}lrrr@{}}
\toprule
\tblhead{Start} & \tblhead{n} & \tblhead{Shift $\rho$} & \tblhead{$\Tpool$} \\
\midrule
Flat & 100 & 0.03 & 0.17 \\
Complement only & 100 & -0.06 & 0.12 \\
Mean field only & 100 & \tblfocus{2.48} & \tblfocus{0.72} \\
\midrule
Full dither & 300 & 2.58 & 0.63 \\
\bottomrule
\end{tabular}
\end{table}

%% file: tables/interventions_main.tex
\begin{table}[tp]
\centering\normalsize
\setlength{\tabcolsep}{3pt}
\papertablestyle
\arrayrulecolor{black}
\caption{Draft interventions on 100 matched confirmation images per pipeline. Photo: photo-style rate. Means retain the flat complement; full dither changes both fields. \tblnote{Bold/underlining rank rendering scores within each pipeline; texture is unranked.}}
\label{tab:interventions}
\begin{tabularx}{\columnwidth}{>{\hsize=1.9\hsize\linewidth=\hsize\raggedright\arraybackslash}X>{\hsize=0.82\hsize\linewidth=\hsize\raggedleft\arraybackslash}X>{\hsize=0.82\hsize\linewidth=\hsize\raggedleft\arraybackslash}X>{\hsize=0.82\hsize\linewidth=\hsize\raggedleft\arraybackslash}X>{\hsize=0.82\hsize\linewidth=\hsize\raggedleft\arraybackslash}X>{\hsize=0.82\hsize\linewidth=\hsize\raggedleft\arraybackslash}X}
\toprule
\tblhead{Input} & \tblhead{$\Tpool$} & \tblhead{Photo} & \tblhead{$S_2$} & \tblhead{$\Delta E$} & \tblhead{Edges} \\
\midrule
\tblgroup{6}{FlowEdit}
Flat & 0.08 & 0.09 & 0.02 & \tblbest{3.1} & \tblbest{0.53} \\
Mean field & 0.50 & \tblsecond{0.26} & \tblsecond{0.05} & \tblsecond{6.0} & \tblsecond{0.49} \\
Full dither & 0.47 & \tblbest{0.79} & \tblbest{0.32} & \tblsecond{6.0} & 0.26 \\
\midrule
\tblgroup{6}{RF-Inversion}
Flat & 0.20 & 0.07 & 0.00 & \tblbest{3.1} & \tblbest{0.37} \\
Mean field & 0.64 & \tblbest{0.82} & \tblbest{0.34} & \tblsecond{4.9} & \tblsecond{0.23} \\
Full dither & 0.62 & \tblsecond{0.81} & \tblsecond{0.30} & 5.3 & 0.15 \\
\bottomrule
\end{tabularx}
\end{table}

%% file: tables/main_rendering.tex
\begin{table*}[t]
\centering\normalsize
\setlength{\tabcolsep}{4pt}
\papertablestyle
\arrayrulecolor{black}
\caption{Photo rendering on the same 300 confirmation images. Methods are grouped by their use of paired draft/photo training. Tile 0.4 is selected on the separate tuning set. Dither + hold and LoRA mean guidance use scale 1.15. Values are point estimates. \tblnote{Bold: best; underline: second-best among the seven methods.}}
\label{tab:main-rendering}
\begin{tabular*}{.98\textwidth}{@{\extracolsep{\fill}}llrrrrrr@{}}
\toprule
\multirow{2}{*}{\tblhead{Family}} & \multirow{2}{*}{\tblhead{Method}} & \multicolumn{2}{c}{\color{black}Overall rendering} & \multicolumn{2}{c}{\color{black}Content recovery} & \multicolumn{2}{c}{\color{black}Regional fidelity} \\
\cmidrule(lr){3-4}\cmidrule(lr){5-6}\cmidrule(l){7-8}
& & \tblhead{$S_2\uparrow$} & \tblhead{HPSv3 $\uparrow$} & \tblhead{Classes $\uparrow$} & \tblhead{Objects $\uparrow$} & \tblhead{$\Delta E\downarrow$} & \tblhead{Edges $\uparrow$} \\
\midrule
\multirow{3}{*}{\shortstack[l]{\textit{Frozen}\\\textit{baselines}}} & SDEdit 0.82 & 0.00 & -5.38 & 0.36 & 0.34 & \tblbest{2.5} & 0.40 \\
 & SDEdit 0.92 & 0.07 & 2.72 & 0.38 & 0.29 & 6.0 & 0.14 \\
 & ControlNet Tile 0.4 & 0.14 & 10.71 & 0.54 & 0.54 & 12.4 & 0.22 \\
\midrule
\multirow{2}{*}{\shortstack[l]{\textit{Training-free}\\\textit{corrections}}} & Dither & 0.35 & \tblsecond{11.35} & 0.51 & 0.41 & 4.1 & 0.19 \\
 & Dither + hold & 0.33 & 9.94 & 0.51 & 0.44 & \tblsecond{3.3} & 0.33 \\
\midrule
\multirow{2}{*}{\shortstack[l]{\textit{Paired-data}\\\textit{adaptation}}} & LoRA alone & \tblsecond{0.38} & 10.93 & \tblbest{0.65} & \tblbest{0.77} & 7.1 & \tblbest{0.44} \\
 & LoRA + mean guidance & \tblbest{0.48} & \tblbest{12.56} & \tblsecond{0.64} & \tblsecond{0.75} & 4.0 & \tblsecond{0.42} \\
\bottomrule
\end{tabular*}
\vspace{-6pt}
\end{table*}

%% file: section/conclusion.tex
\Needspace{5\baselineskip}
\section{Conclusion}\label{sec:conclusion}
Flat drafts alter regional VAE means even when they preserve regional pixel colors. Controlled interventions show that this displacement strongly affects texture generation, while the tested trajectories approximately retain its photo-relative coordinate. The connection explains persistent flatness and motivates dither correction, boundary preservation, and explicit mean guidance. A draft-conditioned LoRA learns a related regional correction and improves content recovery.

\phantomsection\label{sec:limitations}
\noindent\textbf{Limitations.}
Evaluation uses two backbones, fixed masks and schedules, and one LoRA training run. Scores are automated; qualitative cases are selected retrospectively. Shared-direction alignment is empirical. Gaussian retention assumes an autonomous coordinate and ideal flow; coupled generators require endpoint measurements. Guidance convergence assumes isolated updates and a fixed attainable target. LoRA probes jointly vary adaptation and draft-token access. Holding is isolated at scale one on mechanism images; confirmation evaluates scale-and-hold. Supplied partitions and original layouts define the task; recoloring, relocation, and removal remain open settings. Generated detail can alter content, identity, and geometry; the texture metric measures source-relative variation magnitude.

%% file: section/llm_usage.tex
\section*{LLM Usage Statement}
OpenAI ChatGPT 6 Astra assisted with writing the initial manuscript draft, organizing the narrative, and polishing the language of the manuscript. The authors take full responsibility for the final manuscript, code, results, and conclusions.

%% file: appendxi/construction.tex
\section{Construction and Data}\label{app:construction}
All natural images come from COCO val2017~\cite{lin2014coco}, with captions and panoptic annotations~\cite{kirillov2019panoptic}; \cref{tab:data} lists the subsets. The source image and label map receive the same center crop and resize to $512\times512$. Every labeled region is filled with its mean RGB value to form the draft; an unlabeled region remains a separate region in the partition. Eight-bit storage rounds each color component, so equality of regional RGB means is exact before quantization and has rounding error at most $1/(2\cdot255)$ afterward.

The two evaluation subsets require a short image side of at least 420 pixels and no crowd annotation. Each image has two to twelve regions covering at least one percent, an object occupying three to fifty percent, at most ten percent unlabeled pixels, and at most twenty-five segments. A single partition is held fixed across interventions.

\begin{table}[t]
\centering\normalsize\setlength{\tabcolsep}{3pt}
\papertablestyle
\caption{Natural-image subsets. The first five subsets are disjoint. The LoRA subset excludes development, mechanism, confirmation, and reference images, and includes eligible direction-fitting images. The first 100 mechanism images support tuning; the original 48-image LoRA probe and ControlNet selection use subsets of that set. Final LoRA readouts use all 300 confirmation images.}
\label{tab:data}
\begin{tabularx}{\columnwidth}{lrX}
\toprule
\tblhead{Subset} & \tblhead{Images} & \tblhead{Role}\\\midrule
Direction fit & 1,000 & Shared direction and regression\\
Mechanism & 300 & Counterfactual trajectories\\
Confirmation & 300 & Renderer evaluation\\
Photo reference & 1,000 & Distributional distances\\
Development & 48 & Early development\\
LoRA pairs & 2,221 & Paired draft/photo training\\\bottomrule
\end{tabularx}
\end{table}

The synthetic block set contains 5,000 stimuli: fifty base colors crossed with one hundred perturbation conditions. Images have side length 256 pixels. Perturbations include color and gray white noise, band-pass noise, $1/f$ noise, gratings, and natural-texture crops. A uniform color change provides a matched-amplitude control. The tested amplitudes span 0.0003 to 0.08 on the unit image scale. Full-field, half-field, and disk layouts probe sensitivity and spatial leakage. Block statistics use interior latent cells to reduce edge effects. At band scale $b>1$, the perturbation is the unit-standard-deviation difference $G_{b/2}*w-G_b*w$ of periodic Gaussian filters applied to white noise $w$. At $b=1$, it is the normalized residual $w-G_1*w$. The scalar field is shared across RGB channels.

The layered set contains one hundred programmatic layouts with three to twelve flat shapes. Shape areas are drawn on a logarithmic scale between 0.2\% and 50\% of the image, decoupling area from the color and object category structure of natural scenes. Retention is measured across starting noise levels and region-size groups. 

The LoRA pairs are drawn from eligible remaining val2017 images with captions and short side at least 420 pixels. Each target photo is paired with its own mean-color draft and the photo-style caption prompt. Horizontal mirroring is the only recorded augmentation. The training set is a study-specific subset of val2017; COCO's official training split is unused. No mechanism or confirmation image is used to update adapter parameters.

\subsection{Prompt Templates}\label{app:prompts}
The target prompts retain the source caption and specify the desired rendering style. The photo template is: ``RAW photo, 35mm film, photorealistic: \texttt{<caption>}. Real location, real objects, natural light, sharp focus, fine texture.'' The anime template is: ``anime screencap, Japanese animation style, detailed cel shading, clean line art: \texttt{<caption>}.'' The first COCO caption supplies the scene text. The source prompt for text-based editing is: ``a flat vector illustration made of simple solid colour shapes: \texttt{<caption>}''. Diagnostic prompt controls replace the photo appearance request with a flat-illustration request while keeping the latent intervention fixed.

\subsection{Qualitative Selection and Composition}
The introduction teaser uses the staircase scene in column four (counting from one) of the original FLUX contact sheet. Its six complete cells show the source photo, draft, and all four field combinations, with the two off-diagonal interventions highlighted outside the images. Architectural surfaces and railings expose the texture response while changes in the people remain visible. Matrix annotations report geometric texture ratios over all 300 images. Both miniature curves retain all six doses from the 100-image Q1 sweep. No local crop, retouching, or new generation is used.

The cross-method figure uses confirmation index 6 (COCO 166521, television room) for FlowEdit and index 1 (57672, rowers) for RF-Inversion. They show different relationships between means-only and full-dither rendering. The training-free baseline comparison pairs index 18 (245320, skateboard) with index 37 (257896, person in a hat), retaining an example where Tile recovers the subject and the draft corrections reinterpret the silhouette. The appendix retains indices 12 and 3 from the original case set to show further semantic errors.

The LoRA control figure uses confirmation indices 297 (429598, kitchen) and 92 (290833, zebras). It retains all five tested scales, exposing wall, wood, stripe, and grass detail alongside changes in reconstructed geometry. These six main examples were selected retrospectively for the comparisons they illustrate. They serve as case illustrations; population statements use the complete quantitative cohorts and the automated metrics.

The supplementary cases supply lossless comparison sheets with native 512-pixel cells. The manuscript extracts those complete cells without retouching and records every source box in its figure provenance. The four training-free cases form two rows of ten images, with method and group labels below; the other comparisons retain their side labels. No local enlargements or crop-box overlays are added. No image content is generated or enhanced for presentation.

%% file: appendxi/metrics.tex
\section{Metrics and Statistical Conventions}\label{app:metrics}
The metric guide summarizes the meanings and directions of the reported scores (\cref{tab:metric-guide}). The following definitions specify the estimators, eligible regions, calibration thresholds, and resampling units. All style, preference, and content scores are automated measurements.
Performance tables use bold for the best point estimate and underlining for the second-best within the stated comparison group; two-condition comparisons mark only the better value. Ties at displayed precision share the same marking, and all-tied comparisons remain plain. Confidence-interval endpoints retain regular type. Texture and shift magnitudes remain descriptive. Mechanism, selected-setting, and paired-contrast emphasis is specified in the corresponding captions.
\input{tables/metric_guide}
\subsection{Texture and Shading}\label{app:texture}
Texture measures fine luminance variation inside region interiors. Convert the image to grayscale $L\in[0,1]$, form $H_2(L)=L-G_2*L$ with a Gaussian filter of standard deviation two pixels, and erode each source region by five pixels. For an interior $I_r$ with at least 200 pixels, let $t_r(y)=\operatorname{std}_{I_r}(H_2(L_y))$. The pooled statistic is
\begin{equation}\label{eq:texture}
 \overline t(y)=\frac{\sum_{r\in\mathcal I}|I_r|t_r(y)}{\sum_{r\in\mathcal I}|I_r|},\qquad
 T(y)=\frac{\overline t(y)}{\max\{\overline t(x),10^{-6}\}},
\end{equation}
where $\mathcal I$ contains eligible interiors and $x$ is the source photo. Factorial statistics use $\log\max(T,10^{-3})$ and exponentiate the image mean when reporting a geometric mean. The zero point of this scale means no measured interior variation; a value of one matches the photo's pooled magnitude. The ratio measures variation magnitude, including variation caused by artifacts. Shading uses the standard deviation of $G_6*L$ in the same interiors and the same source-relative normalization.

We denote the image-level geometric aggregate by $\Tgeom$:
\begin{equation}
 \Tgeom=\exp\!\left(\frac1n\sum_{i=1}^n\log\max\{T(y_i),10^{-3}\}\right).
\end{equation}
The same geometric aggregation is used for Qwen confirmation and the extended quality tables. The dither-field ablation, cross-method mechanism readouts, and LoRA guidance sweep use the coordinate diagnostic's pooled statistic,
\begin{equation}
 \Tpool=\frac{\sum_{i,r}a_{ir}t_r(y_i)}{\sum_{i,r}a_{ir}t_r(x_i)},
\end{equation}
with full latent-region area $a_{ir}$ as weight, regions covering at least one percent of an image, and at least 200 interior pixels. This statistic pools regional variation before taking the source-relative ratio. Its companion mean shift is the area-weighted $\rho_r(\enc(y_i))$ of re-encoded output images, whereas the causal endpoint study reads terminal latents directly.

The supplementary quality logs originally summarize their image-level texture ratios arithmetically. We recompute $\Tgeom$ from the saved individual scores for all incorporated quality tables. The complete LoRA dose sweep provides the pooled diagnostic's sufficient statistics, so its curve retains $\Tpool$. Each estimator retains its own region eligibility and weighting rules. A ratio above one indicates more measured variation than the source, including possible noise or artifacts.

\subsection{Color and Boundaries}
Regional color error first converts each pixel to CIELAB~\cite{iso2019cielab}, then averages the Lab coordinates within each full region. If $\overline l_r(y)$ is that average and $w_r$ the region area fraction, the score is
\begin{equation}
 \Delta E(y,d)=\sum_r w_r\norm{\overline l_r(y)-\overline l_r(d)}_2.
\end{equation}
This order of operations differs from converting mean RGB to Lab. The threshold of ten in the success measures is a study choice.

Boundary recall uses Canny detections~\cite{canny1986edges} in the generated grayscale image. A draft-boundary pixel is recovered if it lies within four pixels of a detected edge. The reported Edges score subtracts the recall obtained with another image's draft boundaries. This chance-adjusted recall measures alignment with the supplied draft boundaries.

\subsection{Style, Preference, and Content}
The style evaluator uses normalized CLIP ViT-L/14 image and text embeddings~\cite{radford2021clip}. A scaled similarity softmax over five descriptions gives probabilities for photo, anime, flat graphic, three-dimensional render, and painting. The winning category defines the style verdict. This custom zero-shot classifier measures agreement with the five style descriptions. The mechanism study additionally reports the photo-category probability.

HPSv3~\cite{ma2025hpsv3} predicts image--prompt preference using the released model, with its returned mean score used directly. A frozen Mask2Former Swin-L COCO-panoptic checkpoint~\cite{cheng2022mask2former} predicts semantic regions and instances. Classes is the fraction of labeled source pixels assigned their source semantic category. Objects is the fraction of eligible source object instances recovered by a predicted instance of the correct class at intersection-over-union at least 0.5. Eligible objects occupy three to fifty percent of the source image. These are study-defined functions of the frozen segmenter's predictions.

Let $V(y)$ indicate a target-style verdict, $C(y)$ denote Classes, and $H(y)$ denote HPSv3. The success predicates are
\begin{align}
 \Sone(y)&=V(y)\,\mathbf1\{\Delta E(y,d)<10\},\\
 \Stwo(y)&=\Sone(y)\,\mathbf1\{C(y)\geq t_C\}\,
                  \mathbf1\{H(y)\geq t_H\}.
\end{align}
For the photo confirmation set, $t_C=0.54$ and $t_H=7.78$, the rounded tenth percentiles of scores on its source photos. The anime preference threshold is 9.19, calibrated from the evaluated FLUX text-to-image anime reference outputs. The code uses unrounded thresholds. These thresholds are calibrated for the evaluated sets. Qwen uses the same confirmation thresholds.
The fixed unrounded thresholds are $t_C=0.5390432436$, $t_H=7.7805012226$ for photo, and $t_H=9.1918448448$ for anime, shown to ten decimal places. The four gates in $\Stwo$ are target style, color, Classes, and HPSv3; Objects is reported separately. Passing the composite can therefore coexist with missing instances or changed poses.

\subsection{Regional Prediction Readouts}\label{app:probe-metrics}
The numerical photo-relative coordinate in \cref{eq:offset} replaces its denominator by $\max\{\norm{\delta_r}^2,10^{-6}\}$. Encoder energy concentration is $\sum_r\rho_r(z_p)^2/\sum_r\norm{\delta_r}^2$, pooling eligible held-out labeled regions across images without centering the offsets.

For clean-prediction error $e_{0,i}$, let $E_{ir}=(A_ie_{0,i})_r$, $w_{ir}=a_{ir}\mathbf1\{a_{ir}\geq0.01N\}$, and $\langle h\rangle_w=\sum_{i,r}w_{ir}h_{ir}/\sum_{i,r}w_{ir}$. The probe errors are
\begin{align}
 \mathrm{Along}\ u&=\sqrt{\langle\ip{E_{ir}}{u}^2\rangle_w},\\
 \mathrm{Orthogonal}&=\sqrt{\langle\norm{(I_C-uu^\top)E_{ir}}^2\rangle_w},\\
 \mathrm{Within}&=\frac1{nNC}\sum_i\norm{(I_N-P_i)e_{0,i}}_F^2.
\end{align}
The first two are regional RMS norms; Within is a complementary-field MSE. The exact orthogonal loss in \cref{eq:loss-split} uses $\Pi$ and $Q$, with the coupling-aware regional norm derived in \cref{app:operators}. Predicted shift averages $\rho_r(\widehat z_0)$ with the same regional weights. All regional numerators and denominators are pooled again within each bootstrap draw.

\subsection{Distributional Distances}
For features $X=(x_i)_{i=1}^n$ and reference features $Y=(y_j)_{j=1}^m$, the unbiased squared maximum mean discrepancy is
\begin{align}
 \widehat{\mathrm{MMD}}_u^2={}&\frac{\sum_{i\neq i'}k(x_i,x_{i'})}{n(n-1)}
 +\frac{\sum_{j\neq j'}k(y_j,y_{j'})}{m(m-1)}\nonumber\\
 &-\frac{2\sum_{i,j}k(x_i,y_j)}{nm}.
\end{align}
KD uses final DINOv2 ViT-L/14 class-token features~\cite{oquab2024dinov2}, the cubic kernel $k(x,y)=(x^\top y/p+1)^3$ for feature dimension $p$, and a factor of one hundred. It follows the KID estimator~\cite{binkowski2018kid} with a different feature representation. CMMD uses unit-normalized CLIP ViT-L/14 features, a Gaussian kernel with bandwidth ten, and a factor of one thousand, following the CLIP-kernel approach~\cite{jayasumana2024cmmd}. The reference contains 1,000 real photos. Anime distances therefore quantify distance to this photo feature distribution.

\subsection{Aggregation and Intervals}
The factorial interaction reported in \cref{tab:mechanism} is the image-averaged difference of differences, $(\ell_{11}-\ell_{01})-(\ell_{10}-\ell_{00})$, using the log-texture outcomes defined in \cref{eq:effects}.

Image-bootstrap intervals resample complete images, retaining paired outcomes and their regions. Most mean and paired-difference intervals use 2,000 resamples. The nominal-dose retention slope uses 300 image-clustered bootstrap resamples, weighting repeated image draws in the pooled region regression. Encoder energy concentration pools eligible labeled regions of area at least one percent. The endpoint position plot averages eligible regions; per-image texture aggregation follows \cref{eq:texture}. These intervals quantify image-sampling variability with the generation seeds and adapter checkpoint fixed.

The incorporated confirmation estimates use 2,000 resamples with NumPy's default random generator at seed zero. Paired conditions are aligned by COCO image ID and share resampled indices. For $\Tgeom$, the geometric aggregate is recomputed in every draw; for $\Tpool$, its numerator and denominator are recomputed before division. Overlapping 100- and 300-image cohorts share their repeated observations. Legacy mechanism intervals retain their original resampling convention. The matched boundary-holding interval uses Python's random generator with 2,000 paired image resamples at seed zero.

%% file: tables/metric_guide.tex
\begin{table*}[t]
\centering
\normalsize
\setlength{\tabcolsep}{5pt}
\papertablestyle
\caption{How to read the experimental metrics. Arrows indicate the preferred direction for rendering scores. Texture measures detail relative to the source, so it carries no preference arrow. $S_1$ and $S_2$ are success rates defined for this study. Black rules separate comparison groups; numeric emphasis is defined in each table caption.}
\label{tab:metric-guide}
\begin{tabularx}{\textwidth}{p{1.03in}p{3.02in}X}
\arrayrulecolor{black}
\toprule
\tblhead{Metric} & \tblhead{What is measured} & \tblhead{How to interpret it}\\
\midrule
\tblgroup{3}{Appearance diagnostics}
$\Tgeom$ & Geometric mean of image-wise, source-relative interior high-pass variation. & Photo-field, boundary-control, Qwen, and extended quality statistic. \\
$\Tpool$ & Area-weighted regional variation pooled across images, then divided by the source pool. & Dither-field, cross-method, and LoRA control diagnostic. \\
\midrule
\tblgroup{3}{Rendering: fidelity, content, and preference}
$\Delta E\downarrow$ & Area-weighted distance between regional mean Lab colors of output and draft. & Lower: closer regional colors. \\
Edges $\uparrow$ & Draft-boundary recall within four pixels of an output edge, minus mismatched-boundary recall. & Higher: stronger alignment above chance. \\
Classes $\uparrow$ & Fraction of labeled source pixels assigned their source semantic category. & Pixel-level semantic agreement; range $[0,1]$. \\
Objects $\uparrow$ & Fraction of eligible source instances recovered in the correct class at IoU $\geq0.5$. & Instance recovery; range $[0,1]$. \\
HPSv3 $\uparrow$ & Frozen model's predicted image--prompt preference. & An automated preference score. \\
\midrule
\tblgroup{3}{Composite success: fraction of images passing all listed gates}
$\Sone\uparrow$ & Target-style verdict and $\Delta E<10$. & Style and color fidelity. \\
$\Stwo\uparrow$ & $S_1$, Classes $\geq t_C$, and HPSv3 $\geq t_H$. & Adds content and preference requirements. \\
\bottomrule
\end{tabularx}
\vspace{3pt}
\begin{minipage}{\textwidth}
\small
The underlying texture ratio is zero for no measured interior variation and one at the source magnitude; $\Tgeom$ applies a $10^{-3}$ numerical floor. Larger ratios indicate more measured variation, including artifacts. Style is the five-way CLIP target verdict. Calibrated thresholds $t_C,t_H$ are fixed within each target-style evaluation. Brackets accompanying estimates are 95\% image-bootstrap intervals at fixed sampling noise.
\end{minipage}
\end{table*}

%% file: appendxi/implementation.tex
\section{Implementation and Settings}\label{app:implementation}
Both frozen backbones use their native shifted schedules with 28 steps. Standard image-to-image starts are FLUX step 8 ($\sigma\approx0.82$) and Qwen step 7 ($\sigma\approx0.83$), with guidance 3.5 and true classifier-free guidance 4, respectively. Standard rendering uses seed zero. RF-Inversion fixes its target noise at seed one; FlowEdit retains its original fixed stepwise noise sequence. Paired conditions within a method share these choices. The separate fixed-noise LoRA probe uses seed seven. Practical renderers receive the draft, its region partition where needed, and the target prompt; source photos are available only for diagnostic interventions and evaluation.

The comparator protocols include standard and stronger SDEdit-style noising~\cite{meng2022sdedit}, source/target-prompt FlowEdit~\cite{kulikov2025flowedit}, and semantic RF-Inversion~\cite{rout2024rfinversion}. Kontext~\cite{bfl2025kontext} performs instructed editing, Redux~\cite{bflFluxRedux} conditions on the draft image, and InstantX ControlNet-Canny~\cite{zhang2023controlnet} receives region outlines. The original FlowEdit and RF-Inversion baselines use 300 confirmation images; their new flat/means-only/dither comparisons use the first 100. Kontext, Redux, and Canny also use this 100-image subset. Dither amplitude is 0.03, the selected FLUX scale is 1.15, and Qwen uses plain dither. The rank-16 adapter~\cite{hu2021lora} uses early mean guidance during inference as specified below.

The color-layout baseline uses the pretrained Shakker-Labs/FLUX.1-dev-ControlNet-Union-Pro checkpoint~\cite{shakkerControlNetUnionPro}, revision \texttt{b8f5aa82147723080efc9221f4fa5cc030dd38c4}. Tile (mode 1), Blur (3), and Low-quality (6) receive the flat draft directly. Each is tested at strengths 0.4, 0.6, and 0.8 on the first 48 mechanism images. Selection maximizes $S_1$, then minimizes mean color error, maximizes style rate, and prefers strength nearest 0.6. The frozen winners are Tile 0.4, Blur 0.8, and Low-quality 0.6; Tile is the overall winner. All three are evaluated on 100 confirmation photos, and Tile is extended to 300 photos and 100 anime targets without retuning. No ControlNet weights are trained in this study.

\begin{table*}[t]
\centering\small\setlength{\tabcolsep}{4pt}
\papertablestyle
\caption{Implementation settings. The guidance values in RF-Inversion, the generator, and regional mean feedback refer to different operations. Rows describe the executed protocols; the planned LoRA step count is distinguished from the checkpoint used for evaluation.}
\label{tab:settings}
\begin{tabularx}{.96\textwidth}{lX}
\toprule
\tblhead{Component} & \tblhead{Setting}\\\midrule
FLUX.1-dev & 512-pixel square output, 28-step shifted schedule, Euler updates, guidance 3.5, rendering seed 0.\\
Qwen-Image & 512-pixel square output, native 28-step schedule, true CFG 4, empty negative prompt, rendering seed 0.\\
SDEdit-style starts & FLUX step 8 or step 4; Qwen step 7. Start indices count from zero.\\
FlowEdit & Source flat-illustration and target-style prompts; $n_{\max}=24$, source guidance 1.5, target guidance 5.5.\\
RF-Inversion & $\gamma=0.5$, $\eta=0.9$, stop step 7; target noise seed 1 and rendering guidance 3.5.\\
\addlinespace[2pt]
Pretrained interfaces & Kontext guidance 2.5; Redux guidance 3.5; original Canny control strength 0.7; 100-image subset.\\
Color-layout controls & Union Pro Tile, Blur, and Low-quality; guidance 3.5, 28 steps, seed 0, flat draft as control image.\\
\addlinespace[2pt]
Dither & Independent pixel/channel Gaussian noise, standard deviation 0.03; seed from image ID; clip and round to eight bits.\\
Boundary holding & FLUX start step 8, scale 1.15, reset boundary-containing tokens after updates with $8\leq i<16$; no additional model evaluations.\\
\addlinespace[2pt]
LoRA adapter & Rank and scaling parameter 16; attention projections and MLP output projections in 57 blocks; 380 adapted layers, about 56 million trainable parameters.\\
LoRA optimization & AdamW, learning rate $10^{-4}$, weight decay 0.01, batch 4, 100-step warm-up, gradient-norm clipping at 1, gradient checkpointing.\\
LoRA schedule & Planned 2,000 optimization steps; evaluated checkpoint at step 1,500; checkpoints every 250 steps. Training seed 0, guidance input 1.\\
LoRA inference & Pure-noise start, 28 steps, draft tokens, guidance 3.5; mean guidance uses $\gamma=1$, target scale 1.15, first 16 steps.\\
\addlinespace[2pt]
Fixed-noise probe & 300 confirmation images, noise seed 7; pure noise and noised-photo step-8 states; guidance inputs 1 and 3.5. Original training progression uses 48 mechanism images.\\\bottomrule
\end{tabularx}
\end{table*}

The executed settings in \cref{tab:settings} distinguish each model and inference interface. FLUX runs use the supplied diffusers 0.30.1 environment; Qwen and the additional pretrained interfaces use the supplied multi-model environment with diffusers 0.43. Deterministic posterior-location encodings are transformed to each model's native normalized latent units. Both tested VAEs expose sixteen channels at stride eight for the image settings used here. FLUX packs $2\times2$ latent cells into a token covering $16\times16$ image pixels.

Kontext is a resolution exception: the original baseline and dither intervention both receive $512\times512$ inputs and generate $1024\times1024$ outputs under the installed pipeline's default output-area setting. Spatial and latent evaluators resize both to $512\times512$ with Lanczos; HPS uses identical preprocessing within each pair. The other incorporated renderers generate $512\times512$ outputs.

The draft-conditioned adapter concatenates 1,024 encoded draft tokens with 1,024 target image tokens. The conditioning tokens share spatial coordinates with target tokens and use one in the first positional channel to distinguish their source. Only target-token velocities enter the loss. The noise level is drawn inside a randomly selected interval of the inference schedule. Training uses paired photo targets and a photo-style prompt; anime evaluation changes the text at inference.

For transfer tests, the mean-only reference is $z(1,0;\Delta z_d)$ and full dither is $z(1,1;\Delta z_d)=z_d$. Each enters the reference-latent branch of the tested pipeline, retaining that pipeline's original trajectory and prompts. Image-conditioned interfaces are tested separately by replacing their input draft with $\widetilde d_\eta$; these substitutions also change the conditioning representation.

The selected training-free correction uses $z_d+0.15S\delta_d$. For soft masks its actual region-mean displacement is $\delta_d+0.15B\delta_d$, so the scale names an implemented field operation. During boundary holding, the replacement uses the next schedule level and the same start latent and noise. This applies spatial-value enforcement during sampling~\cite{lugmayr2022repaint} to the boundary cost of dither. The method uses one image-to-image trajectory; scaling adds a draft encoding and preserves the denoiser evaluation count. The mean-guided LoRA adjusts a clean prediction, converts it back to a velocity, and continues sampling. These operations intervene at different points in the trajectory.

Approximate per-image times were recorded on the single 96-GB GPU used for the experiments: three seconds for the FLUX training-free pass, four to five seconds for Qwen, and eleven seconds for the draft-token LoRA. A training step at batch four took about 3.5 seconds and 38 GB. Multiplying the executed step count by that time gives roughly 1.46 GPU-hours of optimizer steps; this derived estimate excludes evaluation, caching, loading, and setup. The recorded hardware specification is a single 96-GB GPU, and the compute estimate covers optimizer steps.

The runtime comparison combines different trajectory lengths and token counts. Dither and boundary holding introduce no extra denoiser evaluations relative to their chosen image-to-image pass. The LoRA runs the full schedule from pure noise and doubles the image-token sequence through conditioning. 

%% file: appendxi/additional_results.tex
\section{Additional Results and Controls}\label{app:additional}
The extended results document encoding and trajectory controls, training-free comparisons, and the learned predictor. The complete photo-quality and distribution comparisons appear in \cref{tab:rendering,tab:full-photo}; \cref{fig:failure-examples,fig:failure-camera} show further content failures. Uncertainty for the compact main tables is retained in \cref{tab:mechanism,tab:intervention-intervals,tab:rendering,tab:paired}; probe and dose intervals accompany their full readouts. All incorporated confirmation statistics retain the frozen settings and are recomputed from saved per-image scores or regional sufficient statistics. The matched contrasts in \cref{tab:paired} use common image IDs.

\subsection{Encoding and Trajectory Controls}
\input{tables/mechanism}
\input{tables/controls}
\input{tables/retention}

\begin{figure}[t]
 \centering\includegraphics{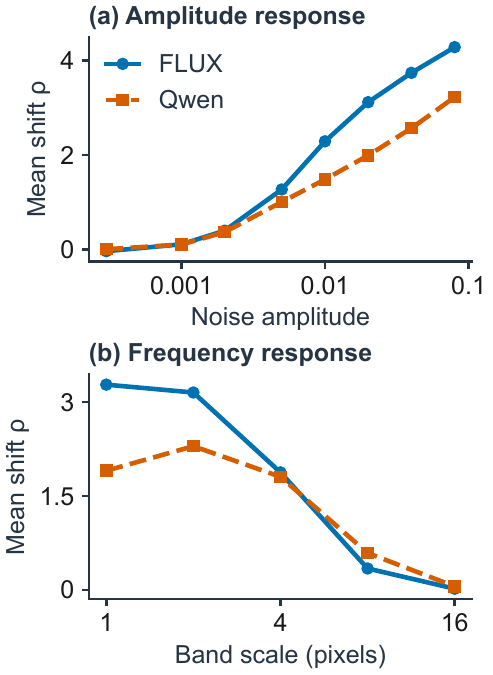}
 \caption{Encoder response in FLUX.1-dev and Qwen-Image. Top: mean displacement along each model's fitted photo direction as white-noise amplitude increases. Bottom: displacement for band-pass perturbations at amplitude 0.02. Points average the tested base colors; lines connect sampled settings and show no uncertainty bands. Fine-scale variation changes a regional latent average.}
 \label{fig:encoder}
\end{figure}
The held-out natural offsets have mean norms 3.29 for FLUX and 2.46 for Qwen in each model's normalized units. The shared-direction photo-versus-draft probe has area under the ROC curve 0.976 and 0.993. At noise amplitude 0.01, the displacement-norm ratio between fine white noise and a uniform color change of equal root-mean-square amplitude is approximately sixteen and thirty-two. These values describe the tested perturbation families.

The half-field experiment distinguishes encoder leakage from the generator's local intervention response. At amplitude 0.01, shift norms in the untouched and textured halves are 1.57 and 2.50 for FLUX, versus 0.07 and 1.68 for Qwen. Thus the two encoders differ in how far a local pixel perturbation affects their region averages. Direct decoding of a shifted flat latent yields fine-variation magnitudes 0.0024 and 0.0047, with mean color changes 0.089 and 0.081 on the unit RGB scale. Direct decoding expresses the shift mainly through color, with fine structure supplied by the subsequent generative trajectory.

The regression features comprise normalized Lab components, their quadratic terms and interactions, and square-root region area. Held-out offset prediction has $R^2=0.73$ for FLUX and $0.38$ for Qwen; a category-only regression gives 0.23 and 0.27. The denominator is squared error relative to the fitting-set mean offset. This supports structured regional dependence beyond the shared direction. These scores quantify held-out offset prediction.

\subsection{Training-Free Comparisons and Boundary Controls}
The following tables collect complete rendering scores, matched interface comparisons, and boundary controls; accompanying notes specify cohorts and interpretation.
\begin{table*}[!tp]
 \tablepanel[\textwidth]{\input{tables/rendering}}
 \par\vspace{12pt}\noindent
 \tablepanel[\columnwidth]{\input{tables/lora_quality_compact}}\hfill
 \tablepanel[\columnwidth]{\input{tables/qwen_compact}}
\end{table*}
\begin{table*}[!tp]
 \tablepanel[\textwidth]{\input{tables/interface_quality}}
 \par\vspace{12pt}\noindent
 \tablepanel[\columnwidth]{\input{tables/intervention_intervals}}\hfill
 \tablepanel[\columnwidth]{\input{tables/controlnet_tuning}}

 \par\vspace{12pt}\noindent
 \begin{minipage}[t]{\columnwidth}
The FlowEdit and RF-Inversion comparisons use all three input conditions on the first 100 confirmation images. Kontext and Redux use flat and dithered image inputs. \Cref{tab:interface-quality} includes the complete quality scores, with geometric texture recomputed from individual quality ratios. The projected shift increases by 0.95 in Redux, while its joint-success rate stays zero; Kontext's mean color error remains above twenty after dithering. Both substitutions change the image-conditioning representation.
 \end{minipage}\hfill
 \begin{minipage}[t]{\columnwidth}
The ControlNet selection evaluates all nine mode/strength pairs on 48 mechanism images (\cref{tab:controlnet-tuning}). Tile 0.4, Blur 0.8, and Low-quality 0.6 are frozen before confirmation scoring. Their matched 100-image photo results appear in \cref{tab:controlnet-matched}; Tile alone is extended to 300 photos and transferred to 100 anime targets (\cref{tab:controlnet-anime}). Tile's higher object recovery and preference than the full draft correction accompany larger color error and weaker boundary alignment in the 300-image photo comparison. Anime uses the photo-selected configuration.
 \end{minipage}
\end{table*}
\begin{table*}[!tp]
 \tablepanel[\textwidth]{\input{tables/controlnet_matched}}
 \par\vspace{12pt}\noindent
 \tablepanel[\columnwidth]{\input{tables/controlnet_anime_compact}}\hfill
 \tablepanel[\columnwidth]{\input{tables/boundary_ablation_compact}}

 \par\vspace{12pt}\noindent
 \begin{minipage}[t]{\columnwidth}
\Cref{tab:boundary-ablation} separates holding from mean-field scaling on the same first 100 mechanism images, with the same prompts, noise, and start step eight. At fixed $\alpha=1$, holding alone increases Edges and reduces color error, while reducing texture. Raising $\alpha$ to 1.15 with holding fixed recovers texture with a smaller cost to both fidelity measures. The table is recomputed from saved per-image scores; its paired interval uses 2,000 image resamples with seed zero. This is a tuning-set ablation, separate from the disjoint 300-image confirmation comparison.
 \end{minipage}\hfill
 \begin{minipage}[t]{\columnwidth}
A further soft-boundary variant uses one-pixel blur, holding to step eighteen, and shift scale 1.25. The full tables include this confirmation run. Its photo-target composite success is close to the original boundary-hold setting, with a persistent content-recovery gap.

The start-noise control changes how much the output retains the supplied mean field. The two-point COCO responses move from 0.79 at an earlier FLUX start to 0.98 at a later start, and from 0.64 to 0.95 for Qwen. Layered-layout responses in \cref{tab:retention} are stratified separately. The region-area predictor explains little of that variation, so its weak fit is retained alongside the strong standard-start dose response.
 \end{minipage}
\end{table*}
\begin{table*}[!tp]
 \tablepanel[\textwidth]{\input{tables/full_photo}}
 \par\vspace{12pt}\noindent
\begin{minipage}[t]{\columnwidth}\vspace{0pt}
 \centering\includegraphics[width=\columnwidth]{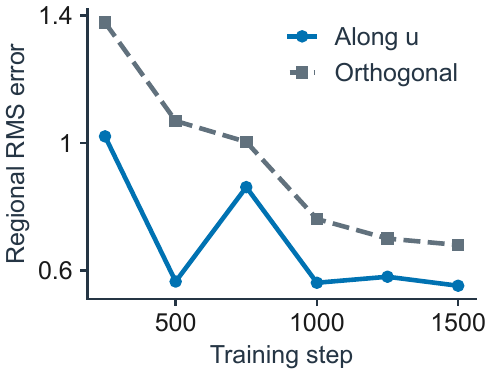}
 \captionof{figure}{Original training progression on 48 mechanism-set probe images, with pure noise, noise seed seven, and guidance input one. Curves join the recorded checkpoints. This development trajectory is separate from the 300-image final-checkpoint confirmation.}
 \label{fig:learning-history}
\end{minipage}\hfill
\begin{minipage}[t]{\columnwidth}\vspace{0pt}
The final-checkpoint probes use all 300 confirmation images and noise seed seven. \Cref{tab:learning} retains both guidance inputs, one and 3.5, and both generated-state conditions. The pure-noise state has $\sigma=1$; the lower-noise diagnostic state is a source photo noised to step eight. The base model receives text without draft tokens; the adapted model additionally receives flat-draft tokens. Probe errors are defined in \cref{app:probe-metrics}. The noised-photo probe at guidance 3.5 increases along-$u$ error from 0.18 to 0.56, showing that the error response depends on the probe state.
\end{minipage}
\end{table*}
\begin{table*}[!tp]
 \tablepanel[\textwidth]{\input{tables/full_anime}}
 \par\vspace{10pt}\noindent
\begin{minipage}[t]{\columnwidth}\vspace{0pt}
 \centering\includegraphics{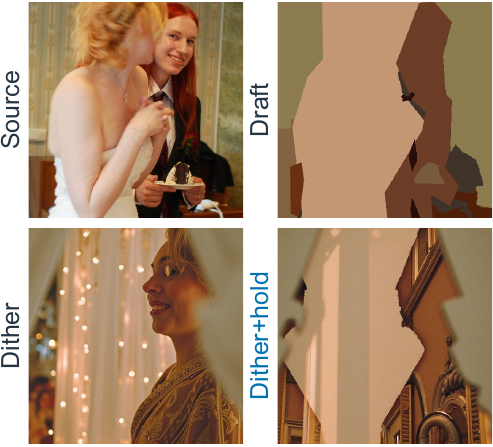}
 \captionof{figure}{Texture recovery with incorrect content. Dither changes the depicted people, and the scaled-dither/holding variant interprets the silhouette as architecture. The four panels are unchanged saved outputs.}
 \label{fig:failure-examples}
\end{minipage}\hfill
\begin{minipage}[t]{\columnwidth}\vspace{0pt}
 \centering\includegraphics{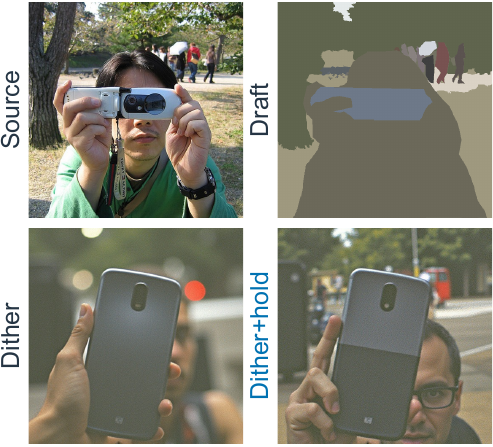}
 \captionof{figure}{A camera becomes a phone after draft rendering. Source, draft, dither, and scaled-dither/holding outputs preserve the full frames. This selected case illustrates a semantic failure.}
 \label{fig:failure-camera}
\end{minipage}
\end{table*}

\subsection{Prediction, Condition, and Guidance Probes}\label{app:probe-conventions}
\begin{table*}[!tp]
 \tablepanel[\textwidth]{\input{tables/learning}}
 \par\vspace{12pt}\noindent
 \tablepanel[\columnwidth]{\input{tables/condition_probe}}\hfill
 \tablepanel[\columnwidth]{\input{tables/lora_readouts}}
\end{table*}
The condition intervention changes only $z_{\rm cond}(\beta)=z_f+\beta S\delta_d$ for $\beta\in\{0,0.5,1,1.5,2\}$ while holding the pure-noise state and prompt fixed (\cref{tab:condition-probe}). The ordinate is the regional weighted mean of $\rho_r(\widehat z_0)$; the reference uses $\rho_r(z_p)$. A fixed additive correction would follow the shifted condition, whereas a photo-level target can respond less to moderate condition doses. The plot uses the injected coefficient $\beta$; its actual regional displacement is $\beta B\delta_d$. The additional full-dither condition uses $z_d$ itself.

All five guidance targets use the same 300 confirmation images, checkpoint, draft tokens, prompt, and noise (\cref{tab:lora-readouts}). Mean guidance acts during the first sixteen steps with gain one. The start comparison fixes the adapter and uses pure noise, flat step-eight initialization, or dithered step-eight initialization without mean guidance. Guidance-dose and initialization experiments therefore vary different parts of the generation process. \Cref{tab:lora-quality} reports the complete independently computed quality scores for the matched 300-image comparison.

\subsection{Paired Comparisons and Content Failures}
\input{tables/paired}
The paired comparisons resample common images to estimate differences between conditions (\cref{tab:paired}). Its contrasts include the 300-image guidance and Tile comparisons and the 100-image full-dither/mean-only comparisons. The qualitative examples in \cref{fig:failure-examples,fig:failure-camera} show semantic errors that persist with increased surface detail.

%% file: tables/mechanism.tex
\begin{table}[tp]
\centering
\normalsize
\setlength{\tabcolsep}{3pt}
\papertablestyle
\arrayrulecolor{black}
\caption{Mechanism measurements. Encoder statistics use held-out regions from 300 photos and the synthetic block set. Half-response amplitude is interpolated in log amplitude at half the response to the largest tested amplitude (0.08). Factorial effects use 300 FLUX images and 100 Qwen images; controls and dose regressions use 100 per model. Brackets are 95\% image-bootstrap intervals. A render is classified as flat when its texture ratio is below 0.3. Values are descriptive. \tblnote{Bold emphasizes encoder concentration, mean-field effects, and retention within each backbone.}}
\label{tab:mechanism}
\begin{tabularx}{\columnwidth}{>{\hsize=1.35\hsize\linewidth=\hsize\raggedright\arraybackslash}X>{\hsize=0.825\hsize\linewidth=\hsize\raggedright\arraybackslash}X>{\hsize=0.825\hsize\linewidth=\hsize\raggedright\arraybackslash}X}
\toprule
\tblhead{Measure} & \tblhead{FLUX} & \tblhead{Qwen} \\
\midrule
Offset energy along u & \tblfocus{0.78} & \tblfocus{0.95} \\
Synthetic/photo direction cosine & 0.93 & 0.95 \\
Half-response noise amplitude & 0.0090 & 0.0119 \\
Mean-field effect on log texture & \shortstack[r]{\tblfocus{2.22}\\{\color{papergray}[2.17,2.28]}} & \shortstack[r]{\tblfocus{1.95}\\{\color{papergray}[1.84,2.06]}} \\
Complement effect on log texture & \shortstack[r]{-0.20\\{\color{papergray}[-0.23,-0.16]}} & \shortstack[r]{-0.20\\{\color{papergray}[-0.27,-0.14]}} \\
Factorial interaction & \shortstack[r]{0.20\\{\color{papergray}[0.14,0.26]}} & \shortstack[r]{0.49\\{\color{papergray}[0.38,0.60]}} \\
Photo complement still flat & 0.91 & 0.94 \\
Random-field log-texture gain & \shortstack[r]{-0.27\\{\color{papergray}[-0.36,-0.17]}} & \shortstack[r]{-0.06\\{\color{papergray}[-0.16,0.05]}} \\
Endpoint slope versus nominal dose & \shortstack[r]{\tblfocus{0.96}\\{\color{papergray}[0.96,0.97]}} & \shortstack[r]{\tblfocus{0.90}\\{\color{papergray}[0.88,0.91]}} \\
\bottomrule
\end{tabularx}
\end{table}

%% file: tables/controls.tex
\begin{table}[tp]
\centering
\normalsize
\setlength{\tabcolsep}{3pt}
\papertablestyle
\arrayrulecolor{black}
\caption{Direction, locality, and prompt controls on 100 mechanism images per model. All texture values are ratios to source-photo variation. Brackets are paired 95\% image-bootstrap intervals for log-texture gains; dashes mark conditions not run.}
\label{tab:controls}
\begin{tabularx}{\columnwidth}{>{\hsize=1.35\hsize\linewidth=\hsize\raggedright\arraybackslash}X>{\hsize=0.825\hsize\linewidth=\hsize\raggedright\arraybackslash}X>{\hsize=0.825\hsize\linewidth=\hsize\raggedright\arraybackslash}X}
\toprule
\tblhead{Control} & \tblhead{FLUX} & \tblhead{Qwen} \\
\midrule
Photo-field log-texture gain & \shortstack[r]{2.09\\{\color{papergray}[1.99,2.20]}} & \shortstack[r]{1.70\\{\color{papergray}[1.59,1.82]}} \\
Random-field log-texture gain & \shortstack[r]{-0.27\\{\color{papergray}[-0.36,-0.17]}} & \shortstack[r]{-0.06\\{\color{papergray}[-0.16,0.05]}} \\
Object-only: object texture & 0.74 & 0.95 \\
Object-only: background texture & 0.24 & 0.29 \\
Shared direction only: texture & 0.88 & 0.84 \\
Shared direction only: color error & 13.5 & 9.0 \\
Orthogonal offset only: texture & 0.20 & -- \\
Photo field + flat prompt: texture & 1.56 & -- \\
\bottomrule
\end{tabularx}
\end{table}

%% file: tables/retention.tex
\begin{table}[tp]
\centering
\normalsize
\setlength{\tabcolsep}{3pt}
\papertablestyle
\arrayrulecolor{black}
\caption{Endpoint retention on 100 synthetic layered layouts, stratified by region size. The median group sizes are 21, 145, and 1,236 latent cells. The logistic fit against the tested detectability predictor has R-squared 0.09 for FLUX and 0.17 for Qwen; an area-only fit gives 0.00 and 0.04. These are endpoint two-point responses, distinct from the multi-dose COCO regression. \tblnote{Bold marks the standard starting noise and its retention readouts.}}
\label{tab:retention}
\begin{tabularx}{\columnwidth}{>{\hsize=1.5\hsize\linewidth=\hsize\raggedright\arraybackslash}X>{\hsize=0.9\hsize\linewidth=\hsize\raggedleft\arraybackslash}X>{\hsize=0.9\hsize\linewidth=\hsize\raggedleft\arraybackslash}X>{\hsize=0.9\hsize\linewidth=\hsize\raggedleft\arraybackslash}X>{\hsize=0.9\hsize\linewidth=\hsize\raggedleft\arraybackslash}X>{\hsize=0.9\hsize\linewidth=\hsize\raggedleft\arraybackslash}X}
\toprule
\tblhead{Model} & \tblhead{Start $s$} & \tblhead{All} & \tblhead{Small} & \tblhead{Mid} & \tblhead{Large} \\
\midrule
FLUX & 0.96 & 0.67 & 0.71 & 0.62 & 0.67 \\
FLUX & 0.92 & 0.86 & 0.88 & 0.82 & 0.88 \\
FLUX & 0.87 & 0.96 & 0.97 & 0.95 & 0.97 \\
FLUX & \tblfocus{0.82} & \tblfocus{1.01} & \tblfocus{1.03} & \tblfocus{1.01} & \tblfocus{0.99} \\
FLUX & 0.71 & 1.01 & 1.04 & 0.99 & 0.99 \\
\midrule
Qwen & 0.98 & 0.17 & 0.22 & 0.16 & 0.14 \\
Qwen & 0.91 & 0.95 & 1.16 & 0.89 & 0.78 \\
Qwen & \tblfocus{0.83} & \tblfocus{1.03} & \tblfocus{1.19} & \tblfocus{0.96} & \tblfocus{0.93} \\
Qwen & 0.71 & 1.03 & 1.12 & 0.98 & 0.98 \\
\bottomrule
\end{tabularx}
\end{table}

%% file: tables/rendering.tex
\begin{table*}[tp]
\centering\normalsize
\setlength{\tabcolsep}{3pt}
\papertablestyle
\arrayrulecolor{black}
\caption{Extended photo-target comparison. Counts refer to the confirmation set; 100-image conditions use its first 100 images. Updated LoRA rows use all 300. Source photos are an evaluation reference. \tblnote{Ranks are within each sample-count group, excluding evaluation references. Bold/underlining mark best/second-best point estimates.}}
\label{tab:rendering}
\begin{tabularx}{.96\textwidth}{>{\hsize=2\hsize\linewidth=\hsize\raggedright\arraybackslash}X>{\hsize=0.35\hsize\linewidth=\hsize\raggedleft\arraybackslash}X>{\hsize=1.85\hsize\linewidth=\hsize\raggedleft\arraybackslash}X>{\hsize=0.8\hsize\linewidth=\hsize\raggedleft\arraybackslash}X>{\hsize=0.8\hsize\linewidth=\hsize\raggedleft\arraybackslash}X>{\hsize=0.8\hsize\linewidth=\hsize\raggedleft\arraybackslash}X>{\hsize=0.7\hsize\linewidth=\hsize\raggedleft\arraybackslash}X>{\hsize=0.7\hsize\linewidth=\hsize\raggedleft\arraybackslash}X}
\toprule
\tblhead{Method} & \tblhead{n} & \tblhead{$S_2$ [95\% CI]} & \tblhead{HPSv3} & \tblhead{Classes} & \tblhead{Objects} & \tblhead{$\Delta E$} & \tblhead{Edges} \\
\midrule
SDEdit 0.82 & 300 & 0.00 [0.00, 0.01] & -5.38 & 0.36 & 0.34 & \tblbest{2.5} & 0.40 \\
SDEdit 0.92 & 300 & 0.07 [0.04, 0.10] & 2.72 & 0.38 & 0.29 & 6.0 & 0.14 \\
FlowEdit & 300 & 0.02 [0.00, 0.03] & -2.28 & 0.42 & 0.45 & 3.0 & \tblbest{0.54} \\
RF-Inversion & 300 & 0.00 [0.00, 0.00] & -2.57 & 0.40 & 0.43 & \tblsecond{2.9} & 0.40 \\
Kontext & 100 & 0.00 [0.00, 0.00] & \tblbest{12.68} & \tblsecond{0.38} & \tblsecond{0.13} & \tblsecond{22.0} & 0.01 \\
Redux & 100 & 0.00 [0.00, 0.00] & -5.93 & 0.27 & 0.11 & \tblbest{9.3} & \tblsecond{0.06} \\
ControlNet-Canny & 100 & 0.00 [0.00, 0.00] & \tblsecond{8.08} & \tblbest{0.42} & \tblbest{0.46} & 24.5 & \tblbest{0.31} \\
Dither & 300 & 0.35 [0.30, 0.41] & \tblsecond{11.35} & 0.51 & 0.41 & 4.1 & 0.19 \\
Dither + hold & 300 & 0.33 [0.28, 0.38] & 9.94 & 0.51 & 0.44 & 3.3 & 0.33 \\
LoRA alone & 300 & \tblsecond{0.38} [0.33, 0.44] & 10.93 & \tblbest{0.65} & \tblbest{0.77} & 7.1 & \tblsecond{0.44} \\
LoRA + mean guidance & 300 & \tblbest{0.48} [0.43, 0.53] & \tblbest{12.56} & \tblsecond{0.64} & \tblsecond{0.75} & 4.0 & 0.42 \\
Source photos & 300 & 0.70 [0.65, 0.75] & 12.34 & 0.82 & 0.91 & 1.1 & 0.37 \\
\bottomrule
\end{tabularx}
\end{table*}

%% file: tables/lora_quality_compact.tex
\begin{table}[tp]
\centering\normalsize
\setlength{\tabcolsep}{3pt}
\papertablestyle
\arrayrulecolor{black}
\caption{Complete quality on the same 300 confirmation images. The final column uses LoRA mean guidance at scale 1.15; the first uses scaled dither with boundary holding. \tblnote{Bold/underlining mark best/second-best rendering scores; texture is unranked.}}
\label{tab:lora-quality}
\begin{tabularx}{\columnwidth}{>{\hsize=1\hsize\linewidth=\hsize\raggedright\arraybackslash}X>{\hsize=1\hsize\linewidth=\hsize\raggedleft\arraybackslash}X>{\hsize=1\hsize\linewidth=\hsize\raggedleft\arraybackslash}X>{\hsize=1\hsize\linewidth=\hsize\raggedleft\arraybackslash}X}
\toprule
\tblhead{Metric} & \tblhead{\shortstack{Dither\\+ hold}} & \tblhead{\shortstack{LoRA\\alone}} & \tblhead{\shortstack{LoRA\\+ guide}} \\
\midrule
$\Tgeom$ & 0.84 & 0.43 & 0.87 \\
Style & \tblbest{0.79} & 0.71 & \tblsecond{0.72} \\
$S_1$ & \tblbest{0.79} & 0.64 & \tblsecond{0.72} \\
$S_2$ & 0.33 & \tblsecond{0.38} & \tblbest{0.48} \\
HPSv3 & 9.94 & \tblsecond{10.93} & \tblbest{12.56} \\
Classes & 0.51 & \tblbest{0.65} & \tblsecond{0.64} \\
Objects & 0.44 & \tblbest{0.77} & \tblsecond{0.75} \\
$\Delta E$ & \tblbest{3.3} & 7.1 & \tblsecond{4.0} \\
Edges & 0.33 & \tblbest{0.44} & \tblsecond{0.42} \\
\bottomrule
\end{tabularx}
\end{table}

%% file: tables/qwen_compact.tex
\begin{table}[tp]
\centering\normalsize
\setlength{\tabcolsep}{3pt}
\papertablestyle
\arrayrulecolor{black}
\caption{Qwen-Image confirmation on 300 images per condition. Only the start image changes. Texture uses geometric aggregation; joint success uses the target-specific style and preference gates. \tblnote{Bold marks the better score within each target-style pair; texture is unranked.}}
\label{tab:qwen}
\begin{tabularx}{\columnwidth}{>{\hsize=1\hsize\linewidth=\hsize\raggedright\arraybackslash}X>{\hsize=1\hsize\linewidth=\hsize\raggedleft\arraybackslash}X>{\hsize=1\hsize\linewidth=\hsize\raggedleft\arraybackslash}X>{\hsize=1\hsize\linewidth=\hsize\raggedleft\arraybackslash}X>{\hsize=1\hsize\linewidth=\hsize\raggedleft\arraybackslash}X}
\toprule
\tblhead{Metric} & \tblhead{\shortstack{Photo\\flat}} & \tblhead{\shortstack{Photo\\dither}} & \tblhead{\shortstack{Anime\\flat}} & \tblhead{\shortstack{Anime\\dither}} \\
\midrule
$\Tgeom$ & 0.18 & 0.55 & 0.27 & 0.72 \\
HPSv3 & -0.4 & \tblbest{9.4} & -3.7 & \tblbest{6.0} \\
$\Delta E$ & \tblbest{6.6} & 8.9 & \tblbest{6.6} & 9.3 \\
Edges & \tblbest{0.21} & 0.07 & \tblbest{0.25} & 0.08 \\
$S_1$ & 0.29 & \tblbest{0.60} & \tblbest{0.85} & 0.57 \\
$S_2$ & 0.02 & \tblbest{0.15} & 0.00 & \tblbest{0.07} \\
\bottomrule
\end{tabularx}
\end{table}

%% file: tables/interface_quality.tex
\begin{table*}[tp]
\centering\normalsize
\setlength{\tabcolsep}{3pt}
\papertablestyle
\arrayrulecolor{black}
\caption{Complete matched input-intervention results, 100 photo-target confirmation images per condition. Texture is the geometric mean of the saved image-level quality ratios; definitions are in \cref{app:metrics}. \tblnote{Ranks are within each pipeline: bold is best; underlining is second-best for three-way comparisons.}}
\label{tab:interface-quality}
\begin{tabularx}{\textwidth}{>{\hsize=3.55\hsize\linewidth=\hsize\raggedright\arraybackslash}X>{\hsize=0.7\hsize\linewidth=\hsize\raggedleft\arraybackslash}X>{\hsize=0.7\hsize\linewidth=\hsize\raggedleft\arraybackslash}X>{\hsize=0.7\hsize\linewidth=\hsize\raggedleft\arraybackslash}X>{\hsize=0.7\hsize\linewidth=\hsize\raggedleft\arraybackslash}X>{\hsize=0.85\hsize\linewidth=\hsize\raggedleft\arraybackslash}X>{\hsize=0.8\hsize\linewidth=\hsize\raggedleft\arraybackslash}X>{\hsize=0.8\hsize\linewidth=\hsize\raggedleft\arraybackslash}X>{\hsize=0.6\hsize\linewidth=\hsize\raggedleft\arraybackslash}X>{\hsize=0.6\hsize\linewidth=\hsize\raggedleft\arraybackslash}X}
\toprule
\tblhead{Method} & \tblhead{$\Tgeom$} & \tblhead{Style} & \tblhead{$S_1$} & \tblhead{$S_2$} & \tblhead{HPSv3} & \tblhead{Classes} & \tblhead{Objects} & \tblhead{$\Delta E$} & \tblhead{Edges} \\
\midrule
FlowEdit & 0.06 & 0.09 & 0.09 & 0.02 & -2.96 & \tblsecond{0.46} & \tblbest{0.44} & \tblbest{3.1} & \tblbest{0.53} \\
FlowEdit / means & 0.54 & \tblsecond{0.26} & \tblsecond{0.24} & \tblsecond{0.05} & \tblsecond{6.47} & \tblbest{0.47} & \tblsecond{0.38} & \tblsecond{6.0} & \tblsecond{0.49} \\
FlowEdit / dither & 0.52 & \tblbest{0.79} & \tblbest{0.74} & \tblbest{0.32} & \tblbest{10.69} & \tblbest{0.47} & 0.31 & \tblsecond{6.0} & 0.26 \\
\midrule
RF-Inversion & 0.21 & 0.07 & 0.07 & 0.00 & -2.57 & \tblsecond{0.45} & \tblbest{0.45} & \tblbest{3.1} & \tblbest{0.37} \\
RF-Inversion / means & 0.71 & \tblbest{0.82} & \tblbest{0.81} & \tblbest{0.34} & \tblsecond{12.13} & \tblbest{0.46} & \tblsecond{0.34} & \tblsecond{4.9} & \tblsecond{0.23} \\
RF-Inversion / dither & 0.69 & \tblsecond{0.81} & \tblsecond{0.80} & \tblsecond{0.30} & \tblbest{12.80} & \tblbest{0.46} & \tblsecond{0.34} & 5.3 & 0.15 \\
\midrule
Kontext & 0.92 & \tblbest{0.32} & 0.00 & 0.00 & \tblbest{12.68} & 0.38 & 0.13 & 22.0 & 0.01 \\
Kontext / dither & 0.90 & 0.28 & 0.00 & 0.00 & 12.31 & \tblbest{0.40} & \tblbest{0.16} & \tblbest{20.1} & \tblbest{0.02} \\
\midrule
Redux & 0.38 & 0.00 & 0.00 & 0.00 & -5.93 & \tblbest{0.27} & 0.11 & \tblbest{9.3} & 0.06 \\
Redux / dither & 0.42 & \tblbest{0.02} & \tblbest{0.01} & 0.00 & \tblbest{-5.76} & 0.25 & \tblbest{0.13} & 9.5 & 0.06 \\
\bottomrule
\end{tabularx}
\end{table*}

%% file: tables/intervention_intervals.tex
\begin{table}[tp]
\centering\normalsize
\setlength{\tabcolsep}{3pt}
\papertablestyle
\arrayrulecolor{black}
\caption{Joint-success uncertainty for the matched 100-image input interventions. Intervals resample images; paired differences are reported separately. \tblnote{Bold/underlining rank point estimates within each pipeline.}}
\label{tab:intervention-intervals}
\begin{tabularx}{\columnwidth}{>{\hsize=0.9\hsize\linewidth=\hsize\raggedright\arraybackslash}X>{\hsize=1.1\hsize\linewidth=\hsize\raggedleft\arraybackslash}X}
\toprule
\tblhead{Input} & \tblhead{$S_2$ [95\% CI]} \\
\midrule
\tblgroup{2}{FlowEdit}
Flat & 0.02 [0.00, 0.05] \\
Mean field & \tblsecond{0.05} [0.01, 0.10] \\
Full dither & \tblbest{0.32} [0.23, 0.41] \\
\midrule
\tblgroup{2}{RF-Inversion}
Flat & 0.00 [0.00, 0.00] \\
Mean field & \tblbest{0.34} [0.25, 0.43] \\
Full dither & \tblsecond{0.30} [0.21, 0.39] \\
\bottomrule
\end{tabularx}
\end{table}

%% file: tables/controlnet_tuning.tex
\begin{table}[tp]
\centering\normalsize
\setlength{\tabcolsep}{2pt}
\papertablestyle
\arrayrulecolor{black}
\caption{All nine ControlNet tuning settings on 48 mechanism images. Selection maximizes $S_1$, then minimizes color error, maximizes style rate, and prefers strength nearest 0.6. Tile 0.4 is the frozen overall winner. \tblnote{Bold marks the selected setting within each mode.}}
\label{tab:controlnet-tuning}
\begin{tabularx}{\columnwidth}{>{\hsize=1.4\hsize\linewidth=\hsize\raggedright\arraybackslash}X>{\hsize=0.8\hsize\linewidth=\hsize\raggedleft\arraybackslash}X>{\hsize=0.8\hsize\linewidth=\hsize\raggedleft\arraybackslash}X>{\hsize=1\hsize\linewidth=\hsize\raggedleft\arraybackslash}X>{\hsize=1\hsize\linewidth=\hsize\raggedleft\arraybackslash}X}
\toprule
\tblhead{Mode} & \tblhead{Scale} & \tblhead{$S_1\uparrow$} & \tblhead{$\Delta E\downarrow$} & \tblhead{Photo} \\
\midrule
Tile & \tblfocus{0.4} & \tblfocus{0.19} & \tblfocus{13.54} & \tblfocus{0.69} \\
Tile & 0.6 & 0.04 & 9.19 & 0.04 \\
Tile & 0.8 & 0.00 & 9.92 & 0.00 \\
\midrule
Blur & 0.4 & 0.02 & 16.32 & 0.58 \\
Blur & 0.6 & 0.00 & 10.44 & 0.06 \\
Blur & \tblfocus{0.8} & \tblfocus{0.02} & \tblfocus{10.39} & \tblfocus{0.02} \\
\midrule
Low-quality & 0.4 & 0.04 & 19.12 & 0.81 \\
Low-quality & \tblfocus{0.6} & \tblfocus{0.15} & \tblfocus{13.14} & \tblfocus{0.67} \\
Low-quality & 0.8 & 0.02 & 10.98 & 0.08 \\
\bottomrule
\end{tabularx}
\end{table}

%% file: tables/controlnet_matched.tex
\begin{table*}[tp]
\centering\normalsize
\setlength{\tabcolsep}{3pt}
\papertablestyle
\arrayrulecolor{black}
\caption{Color-layout controls and draft corrections on the same 100 photo-target confirmation images. Each control mode uses its tuning-selected strength. Texture is the geometric mean of the saved image-level quality ratios; definitions are in \cref{app:metrics}. \tblnote{Bold/underlining mark best/second-best rendering scores; texture is unranked.}}
\label{tab:controlnet-matched}
\begin{tabularx}{\textwidth}{>{\hsize=3.55\hsize\linewidth=\hsize\raggedright\arraybackslash}X>{\hsize=0.7\hsize\linewidth=\hsize\raggedleft\arraybackslash}X>{\hsize=0.7\hsize\linewidth=\hsize\raggedleft\arraybackslash}X>{\hsize=0.7\hsize\linewidth=\hsize\raggedleft\arraybackslash}X>{\hsize=0.7\hsize\linewidth=\hsize\raggedleft\arraybackslash}X>{\hsize=0.85\hsize\linewidth=\hsize\raggedleft\arraybackslash}X>{\hsize=0.8\hsize\linewidth=\hsize\raggedleft\arraybackslash}X>{\hsize=0.8\hsize\linewidth=\hsize\raggedleft\arraybackslash}X>{\hsize=0.6\hsize\linewidth=\hsize\raggedleft\arraybackslash}X>{\hsize=0.6\hsize\linewidth=\hsize\raggedleft\arraybackslash}X}
\toprule
\tblhead{Method} & \tblhead{$\Tgeom$} & \tblhead{Style} & \tblhead{$S_1$} & \tblhead{$S_2$} & \tblhead{HPSv3} & \tblhead{Classes} & \tblhead{Objects} & \tblhead{$\Delta E$} & \tblhead{Edges} \\
\midrule
ControlNet Canny & 0.44 & \tblsecond{0.83} & 0.00 & 0.00 & 8.08 & 0.42 & 0.46 & 24.5 & 0.31 \\
ControlNet Tile 0.4 & 0.49 & 0.68 & 0.24 & 0.12 & \tblsecond{10.67} & \tblbest{0.55} & \tblbest{0.56} & 12.5 & 0.21 \\
ControlNet Blur 0.8 & 0.05 & 0.02 & 0.01 & 0.00 & -8.57 & 0.32 & 0.30 & 10.0 & \tblbest{0.56} \\
ControlNet Low-quality 0.6 & 0.42 & 0.68 & 0.23 & 0.12 & 7.77 & \tblsecond{0.53} & \tblsecond{0.53} & 12.0 & 0.29 \\
SDEdit 0.82 & 0.17 & 0.01 & 0.01 & 0.00 & -5.32 & 0.39 & 0.32 & \tblbest{2.6} & \tblsecond{0.37} \\
Dither & 0.67 & 0.80 & \tblsecond{0.80} & \tblsecond{0.32} & \tblbest{11.63} & 0.49 & 0.37 & 4.4 & 0.17 \\
Dither + hold & 0.79 & \tblbest{0.87} & \tblbest{0.87} & \tblbest{0.35} & 9.98 & 0.49 & 0.38 & \tblsecond{3.5} & 0.30 \\
\bottomrule
\end{tabularx}
\end{table*}

%% file: tables/controlnet_anime_compact.tex
\begin{table}[tp]
\centering\normalsize
\setlength{\tabcolsep}{3pt}
\papertablestyle
\arrayrulecolor{black}
\caption{Anime-target transfer on the same 100 images. Tile retains its photo-tuned strength 0.4; all texture values use geometric aggregation. \tblnote{Bold/underlining mark best/second-best rendering scores; texture is unranked.}}
\label{tab:controlnet-anime}
\begin{tabularx}{\columnwidth}{>{\hsize=1\hsize\linewidth=\hsize\raggedright\arraybackslash}X>{\hsize=1\hsize\linewidth=\hsize\raggedleft\arraybackslash}X>{\hsize=1\hsize\linewidth=\hsize\raggedleft\arraybackslash}X>{\hsize=1\hsize\linewidth=\hsize\raggedleft\arraybackslash}X>{\hsize=1\hsize\linewidth=\hsize\raggedleft\arraybackslash}X}
\toprule
\tblhead{Metric} & \tblhead{\shortstack{Tile\\0.4}} & \tblhead{\shortstack{SDEdit\\0.82}} & \tblhead{Dither} & \tblhead{\shortstack{Dither\\+ hold}} \\
\midrule
$\Tgeom$ & 0.92 & 0.19 & 0.82 & 0.89 \\
Style & \tblbest{0.99} & 0.53 & 0.93 & \tblsecond{0.95} \\
$S_1$ & 0.21 & 0.53 & \tblsecond{0.93} & \tblbest{0.95} \\
$S_2$ & 0.05 & 0.00 & \tblbest{0.19} & \tblsecond{0.13} \\
HPSv3 & \tblsecond{7.17} & -6.41 & \tblbest{8.12} & 5.84 \\
Classes & \tblbest{0.49} & 0.40 & 0.42 & \tblsecond{0.44} \\
Objects & \tblbest{0.51} & 0.35 & \tblsecond{0.37} & \tblsecond{0.37} \\
$\Delta E$ & 14.4 & \tblbest{2.7} & 4.4 & \tblsecond{3.6} \\
Edges & 0.27 & \tblbest{0.37} & 0.16 & \tblsecond{0.29} \\
\bottomrule
\end{tabularx}
\end{table}

%% file: tables/boundary_ablation_compact.tex
\begin{table}[tp]
\centering\normalsize
\setlength{\tabcolsep}{3pt}
\papertablestyle
\arrayrulecolor{black}
\caption{Matched boundary-holding and scale controls on 100 mechanism images at start step eight. At fixed scale one, holding raises Edges by 0.15 [0.13, 0.17] (paired 95\% image-bootstrap interval). \tblnote{Bold/underlining rank fidelity scores; texture remains descriptive.}}
\label{tab:boundary-ablation}
\begin{tabularx}{\columnwidth}{>{\hsize=1.65\hsize\linewidth=\hsize\raggedright\arraybackslash}X>{\hsize=0.65\hsize\linewidth=\hsize\raggedleft\arraybackslash}X>{\hsize=0.9\hsize\linewidth=\hsize\raggedleft\arraybackslash}X>{\hsize=0.9\hsize\linewidth=\hsize\raggedleft\arraybackslash}X>{\hsize=0.9\hsize\linewidth=\hsize\raggedleft\arraybackslash}X}
\toprule
\tblhead{Hold} & \tblhead{$\alpha$} & \tblhead{$\Tgeom$} & \tblhead{$\Delta E$} & \tblhead{Edges} \\
\midrule
Off & 1.00 & 0.68 & 4.3 & 0.18 \\
To step 16 & 1.00 & 0.58 & \tblbest{3.2} & \tblbest{0.33} \\
To step 16 & 1.15 & 0.82 & \tblsecond{3.5} & \tblsecond{0.31} \\
\bottomrule
\end{tabularx}
\end{table}

%% file: tables/full_photo.tex
\begin{table*}[tp]
\centering\normalsize
\setlength{\tabcolsep}{3pt}
\papertablestyle
\arrayrulecolor{black}
\caption{Photo-target appearance and distribution scores. LoRA quality uses 300 images. Expanded unguided LoRA has no KD estimate; other KD scores retain their original cohorts. \tblnote{Bold/underlining: best/second-best within each sample-count group, excluding references. KD is unranked across its different cohorts.}}
\label{tab:full-photo}
\begin{tabularx}{.94\textwidth}{>{\hsize=2\hsize\linewidth=\hsize\raggedright\arraybackslash}X>{\hsize=0.4\hsize\linewidth=\hsize\raggedleft\arraybackslash}X>{\hsize=1.8\hsize\linewidth=\hsize\raggedleft\arraybackslash}X>{\hsize=0.65\hsize\linewidth=\hsize\raggedleft\arraybackslash}X>{\hsize=0.65\hsize\linewidth=\hsize\raggedleft\arraybackslash}X>{\hsize=0.8\hsize\linewidth=\hsize\raggedleft\arraybackslash}X>{\hsize=0.7\hsize\linewidth=\hsize\raggedleft\arraybackslash}X}
\toprule
\tblhead{Method} & \tblhead{n} & \tblhead{$S_2$ [95\% CI]} & \tblhead{$S_1$} & \tblhead{Style} & \tblhead{HPSv3} & \tblhead{KD} \\
\midrule
Source photos & 300 & 0.70 [0.65, 0.75] & 0.85 & 0.85 & 12.34 & 0.8 \\
Flat drafts & 300 & 0.00 [0.00, 0.00] & 0.00 & 0.00 & -8.58 & 319.9 \\
Text-to-image & 100 & 0.00 [0.00, 0.00] & 0.00 & 0.55 & 14.63 & 20.4 \\
SDEdit 0.82 & 300 & 0.00 [0.00, 0.01] & 0.03 & 0.03 & -5.38 & 163.5 \\
SDEdit 0.92 & 300 & 0.07 [0.04, 0.10] & 0.43 & 0.45 & 2.72 & 36.9 \\
FlowEdit & 300 & 0.02 [0.00, 0.03] & 0.09 & 0.09 & -2.28 & 118.1 \\
RF-Inversion & 300 & 0.00 [0.00, 0.00] & 0.08 & 0.08 & -2.57 & 119.0 \\
Kontext & 100 & 0.00 [0.00, 0.00] & 0.00 & \tblsecond{0.32} & \tblbest{12.68} & 29.4 \\
Redux & 100 & 0.00 [0.00, 0.00] & 0.00 & 0.00 & -5.93 & 244.0 \\
ControlNet-Canny & 100 & 0.00 [0.00, 0.00] & 0.00 & \tblbest{0.83} & \tblsecond{8.08} & 14.5 \\
Dither & 300 & 0.35 [0.30, 0.41] & \tblbest{0.80} & \tblbest{0.80} & \tblsecond{11.35} & 8.2 \\
Dither + hold & 300 & 0.33 [0.28, 0.38] & \tblsecond{0.79} & \tblsecond{0.79} & 9.94 & 6.7 \\
Soft boundary variant & 300 & 0.33 [0.27, 0.38] & \tblsecond{0.79} & \tblsecond{0.79} & 9.27 & 7.2 \\
LoRA alone & 300 & \tblsecond{0.38} [0.33, 0.44] & 0.64 & 0.71 & 10.93 & -- \\
LoRA + mean guidance & 300 & \tblbest{0.48} [0.43, 0.53] & 0.72 & 0.72 & \tblbest{12.56} & 6.8 \\
\bottomrule
\end{tabularx}
\end{table*}

%% file: tables/full_anime.tex
\begin{table*}[tp]
\centering
\normalsize
\setlength{\tabcolsep}{3pt}
\papertablestyle
\arrayrulecolor{black}
\caption{Anime-target appearance and distribution scores on confirmation images; counts vary by interface. $S_1,S_2$: defined composites. Style: five-way classifier verdict rate. KD compares DINOv2 features with real photos using the scaled cubic-kernel estimator, also for anime targets. Brackets: 95\% image-bootstrap intervals. \tblnote{Bold/underlining: best/second-best within each sample-count group, excluding references. KD measures distance to the photo reference distribution.}}
\label{tab:full-anime}
\begin{tabularx}{.95\textwidth}{>{\hsize=2\hsize\linewidth=\hsize\raggedright\arraybackslash}X>{\hsize=0.4\hsize\linewidth=\hsize\raggedleft\arraybackslash}X>{\hsize=1.7\hsize\linewidth=\hsize\raggedright\arraybackslash}X>{\hsize=0.75\hsize\linewidth=\hsize\raggedleft\arraybackslash}X>{\hsize=0.75\hsize\linewidth=\hsize\raggedleft\arraybackslash}X>{\hsize=0.75\hsize\linewidth=\hsize\raggedleft\arraybackslash}X>{\hsize=0.65\hsize\linewidth=\hsize\raggedleft\arraybackslash}X}
\toprule
\tblhead{Method} & \tblhead{n} & \tblhead{$S_2\uparrow$ [95\% CI]} & \tblhead{$S_1\uparrow$} & \tblhead{Style $\uparrow$} & \tblhead{HPSv3 $\uparrow$} & \tblhead{KD} \\
\midrule
\tblgroup{7}{Evaluation references}
Source photos & 300 & 0.01 [0.00, 0.02] & 0.01 & 0.01 & 8.88 & 0.8 \\
Flat drafts & 300 & 0.00 [0.00, 0.00] & 0.03 & 0.03 & -9.84 & 319.9 \\
Text-to-image & 100 & 0.00 [0.00, 0.00] & 0.02 & 0.97 & 11.91 & 94.0 \\
\midrule
\tblgroup{7}{Frozen baselines and pretrained interfaces}
SDEdit 0.82 & 300 & 0.00 [0.00, 0.00] & 0.51 & 0.51 & -6.69 & 176.9 \\
SDEdit 0.92 & 300 & 0.04 [0.02, 0.06] & 0.90 & \tblsecond{0.95} & 2.48 & 119.6 \\
FlowEdit & 300 & 0.02 [0.01, 0.04] & 0.50 & 0.50 & -3.40 & 150.6 \\
RF-Inversion & 300 & 0.01 [0.00, 0.02] & 0.75 & 0.75 & -4.09 & 158.9 \\
Kontext & 100 & \tblsecond{0.01} [0.00, 0.03] & 0.01 & \tblsecond{0.90} & \tblbest{10.77} & 125.9 \\
Redux & 100 & 0.00 [0.00, 0.00] & \tblsecond{0.35} & 0.61 & -7.49 & 241.2 \\
ControlNet-Canny & 100 & 0.00 [0.00, 0.00] & 0.00 & \tblbest{0.97} & 5.46 & 66.1 \\
\midrule
\tblgroup{7}{Draft corrections}
Dither & 300 & \tblbest{0.21} [0.17, 0.26] & \tblbest{0.96} & \tblbest{0.96} & \tblbest{7.07} & 79.9 \\
Dither + hold & 300 & \tblsecond{0.14} [0.11, 0.19] & 0.94 & 0.94 & \tblsecond{4.96} & 84.7 \\
Soft boundary variant & 300 & 0.13 [0.10, 0.17] & \tblsecond{0.95} & \tblsecond{0.95} & 3.46 & 92.5 \\
\midrule
\tblgroup{7}{Paired-data adaptation}
LoRA + mean guidance & 100 & \tblbest{0.32} [0.23, 0.41] & \tblbest{0.81} & 0.81 & \tblsecond{9.50} & 76.5 \\
\bottomrule
\end{tabularx}
\end{table*}

%% file: tables/learning.tex
\begin{table*}[tp]
\centering\normalsize
\setlength{\tabcolsep}{3pt}
\papertablestyle
\arrayrulecolor{black}
\caption{Clean-prediction probes on 300 confirmation images, fixed noise seed seven. Base is text-conditioned without draft tokens; LoRA also receives flat-draft tokens. The noised-photo state is at step eight. Along-$u$ and orthogonal errors are regional RMS norms; Within is the implemented complementary-field MSE. \tblnote{Bold marks the lower error within each fixed-state, fixed-guidance pair; shift is unranked.}}
\label{tab:learning}
\begin{tabularx}{\textwidth}{>{\hsize=1.4\hsize\linewidth=\hsize\raggedright\arraybackslash}X>{\hsize=1.05\hsize\linewidth=\hsize\raggedleft\arraybackslash}X>{\hsize=1.05\hsize\linewidth=\hsize\raggedleft\arraybackslash}X>{\hsize=1.05\hsize\linewidth=\hsize\raggedleft\arraybackslash}X>{\hsize=0.45\hsize\linewidth=\hsize\raggedleft\arraybackslash}X}
\toprule
\tblhead{State / guidance / model} & \tblhead{Shift $\rho$ [95\% CI]} & \tblhead{Along $u$ [95\% CI]} & \tblhead{Orthogonal [95\% CI]} & \tblhead{Within MSE} \\
\midrule
Pure noise / 1.0 / Base & 2.27 [2.19, 2.35] & 0.87 [0.83, 0.91] & 2.06 [1.97, 2.15] & 0.337 \\
Pure noise / 1.0 / LoRA & 2.63 [2.58, 2.68] & \tblbest{0.53} [0.51, 0.56] & \tblbest{0.71} [0.67, 0.74] & \tblbest{0.310} \\
\midrule
Pure noise / 3.5 / Base & 2.31 [2.20, 2.43] & 1.16 [1.09, 1.22] & 3.16 [3.01, 3.31] & 0.496 \\
Pure noise / 3.5 / LoRA & 2.63 [2.58, 2.68] & \tblbest{0.57} [0.54, 0.60] & \tblbest{0.92} [0.87, 0.97] & \tblbest{0.409} \\
\midrule
Noised photo / 1.0 / Base & 2.87 [2.80, 2.95] & \tblbest{0.13} [0.12, 0.13] & 0.34 [0.33, 0.35] & 0.206 \\
Noised photo / 1.0 / LoRA & 2.81 [2.75, 2.88] & 0.15 [0.14, 0.16] & \tblbest{0.29} [0.28, 0.29] & \tblbest{0.194} \\
\midrule
Noised photo / 3.5 / Base & 2.86 [2.79, 2.93] & \tblbest{0.18} [0.17, 0.18] & 0.45 [0.44, 0.47] & 0.274 \\
Noised photo / 3.5 / LoRA & 2.36 [2.30, 2.43] & 0.56 [0.55, 0.57] & \tblbest{0.44} [0.43, 0.45] & \tblbest{0.237} \\
\bottomrule
\end{tabularx}
\end{table*}

%% file: tables/condition_probe.tex
\begin{table}[tp]
\centering\normalsize
\setlength{\tabcolsep}{3pt}
\papertablestyle
\arrayrulecolor{black}
\caption{Condition-token probes on the same 300 images. Numeric rows use $z_f+\beta S\delta_d$; full dither uses $z_d$. The source-photo reference is 2.89. All generated states start from pure noise.}
\label{tab:condition-probe}
\begin{tabularx}{.96\columnwidth}{>{\hsize=0.75\hsize\linewidth=\hsize\raggedright\arraybackslash}X>{\hsize=0.75\hsize\linewidth=\hsize\raggedleft\arraybackslash}X>{\hsize=1.5\hsize\linewidth=\hsize\raggedleft\arraybackslash}X}
\toprule
\tblhead{Guidance input} & \tblhead{Injected $\beta$} & \tblhead{Predicted $\rho$ [95\% CI]} \\
\midrule
1.0 & 0.0 & 2.63 [2.58, 2.68] \\
1.0 & 0.5 & 2.79 [2.74, 2.84] \\
1.0 & 1.0 & 2.93 [2.88, 2.99] \\
1.0 & 1.5 & 4.04 [3.97, 4.12] \\
1.0 & 2.0 & 5.86 [5.76, 5.95] \\
1.0 & Full dither & 2.92 [2.86, 2.97] \\
\midrule
3.5 & 0.0 & 2.63 [2.58, 2.68] \\
3.5 & 0.5 & 2.83 [2.77, 2.89] \\
3.5 & 1.0 & 3.17 [3.10, 3.23] \\
3.5 & 1.5 & 3.86 [3.78, 3.94] \\
3.5 & 2.0 & 5.70 [5.60, 5.81] \\
3.5 & Full dither & 3.14 [3.08, 3.21] \\
\bottomrule
\end{tabularx}
\end{table}

%% file: tables/lora_readouts.tex
\begin{table}[tp]
\centering\normalsize
\setlength{\tabcolsep}{3pt}
\papertablestyle
\arrayrulecolor{black}
\caption{Regional appearance readouts on the same 300 confirmation images. Guidance sweeps share the checkpoint, draft tokens, prompt, and noise; flat and dither image-to-image starts share step eight. Statistics use re-encoded finished images. Brackets show 95\% image-bootstrap intervals. \tblnote{Bold identifies the selected target scale 1.15.}}
\label{tab:lora-readouts}
\begin{tabularx}{\columnwidth}{>{\hsize=1.2\hsize\linewidth=\hsize\raggedright\arraybackslash}X>{\hsize=0.9\hsize\linewidth=\hsize\raggedleft\arraybackslash}X>{\hsize=0.9\hsize\linewidth=\hsize\raggedleft\arraybackslash}X}
\toprule
\tblhead{Condition} & \tblhead{Shift $\rho$} & \tblhead{$\Tpool$} \\
\midrule
source photo & \shortstack[r]{2.89\\{\color{papergray}[2.82, 2.96]}} & \shortstack[r]{1.00\\{\color{papergray}[1.00, 1.00]}} \\
flat draft & \shortstack[r]{0.00\\{\color{papergray}[0.00, 0.00]}} & \shortstack[r]{0.01\\{\color{papergray}[0.00, 0.01]}} \\
dithered draft & \shortstack[r]{2.54\\{\color{papergray}[2.49, 2.59]}} & \shortstack[r]{0.38\\{\color{papergray}[0.36, 0.40]}} \\
\midrule
LoRA noise start & \shortstack[r]{1.45\\{\color{papergray}[1.41, 1.50]}} & \shortstack[r]{0.40\\{\color{papergray}[0.38, 0.42]}} \\
LoRA flat start & \shortstack[r]{0.09\\{\color{papergray}[0.07, 0.10]}} & \shortstack[r]{0.08\\{\color{papergray}[0.07, 0.08]}} \\
LoRA dither start & \shortstack[r]{1.96\\{\color{papergray}[1.91, 2.00]}} & \shortstack[r]{0.35\\{\color{papergray}[0.34, 0.37]}} \\
\midrule
LoRA guidance x0 & \shortstack[r]{0.04\\{\color{papergray}[0.03, 0.06]}} & \shortstack[r]{0.13\\{\color{papergray}[0.12, 0.14]}} \\
LoRA guidance x0.5 & \shortstack[r]{1.12\\{\color{papergray}[1.10, 1.15]}} & \shortstack[r]{0.29\\{\color{papergray}[0.27, 0.30]}} \\
LoRA guidance x1 & \shortstack[r]{2.36\\{\color{papergray}[2.31, 2.41]}} & \shortstack[r]{0.59\\{\color{papergray}[0.56, 0.63]}} \\
LoRA guidance x1.15 & \shortstack[r]{\tblfocus{2.74}\\{\color{papergray}[2.68, 2.80]}} & \shortstack[r]{\tblfocus{0.78}\\{\color{papergray}[0.74, 0.82]}} \\
LoRA guidance x1.3 & \shortstack[r]{3.11\\{\color{papergray}[3.04, 3.17]}} & \shortstack[r]{1.09\\{\color{papergray}[1.03, 1.15]}} \\
\bottomrule
\end{tabularx}
\end{table}

%% file: tables/paired.tex
\begin{table*}[tp]
\centering\normalsize
\setlength{\tabcolsep}{3pt}
\papertablestyle
\arrayrulecolor{black}
\caption{Paired confirmation contrasts. Full denotes scaled dither with boundary holding; means denotes the dither-induced mean field. Each contrast resamples the same image IDs in its two conditions. \tblnote{Bold marks differences whose displayed intervals exclude zero, in either direction.}}
\label{tab:paired}
\begin{tabularx}{.84\textwidth}{>{\hsize=1.6\hsize\linewidth=\hsize\raggedright\arraybackslash}X>{\hsize=0.3\hsize\linewidth=\hsize\raggedleft\arraybackslash}X>{\hsize=0.5\hsize\linewidth=\hsize\raggedleft\arraybackslash}X>{\hsize=1.6\hsize\linewidth=\hsize\raggedleft\arraybackslash}X}
\toprule
\tblhead{Contrast} & \tblhead{n} & \tblhead{Outcome} & \tblhead{Difference [95\% CI]} \\
\midrule
Guided - alone & 300 & $S_2$ & \tblfocus{0.10} [0.04, 0.15] \\
Guided - alone & 300 & $\Delta E$ & \tblfocus{-3.15} [-3.40, -2.90] \\
Guided - alone & 300 & Edges & \tblfocus{-0.02} [-0.02, -0.01] \\
Guided - alone & 300 & HPSv3 & \tblfocus{1.63} [1.39, 1.89] \\
Guided - alone & 300 & Objects & -0.02 [-0.05, 0.01] \\
\midrule
Guided - full & 300 & $S_2$ & \tblfocus{0.15} [0.09, 0.20] \\
Guided - full & 300 & $\Delta E$ & \tblfocus{0.63} [0.45, 0.82] \\
Guided - full & 300 & Edges & \tblfocus{0.09} [0.07, 0.11] \\
Guided - full & 300 & HPSv3 & \tblfocus{2.62} [2.09, 3.17] \\
Guided - full & 300 & Objects & \tblfocus{0.31} [0.26, 0.36] \\
\midrule
Tile - full & 300 & $S_2$ & \tblfocus{-0.20} [-0.25, -0.14] \\
Tile - full & 300 & $\Delta E$ & \tblfocus{9.03} [8.47, 9.61] \\
Tile - full & 300 & Edges & \tblfocus{-0.11} [-0.14, -0.09] \\
Tile - full & 300 & HPSv3 & \tblfocus{0.77} [0.18, 1.38] \\
Tile - full & 300 & Objects & \tblfocus{0.10} [0.05, 0.15] \\
\midrule
Full - dither & 300 & $S_2$ & -0.02 [-0.05, 0.02] \\
Full - dither & 300 & $\Delta E$ & \tblfocus{-0.79} [-0.88, -0.71] \\
Full - dither & 300 & Edges & \tblfocus{0.14} [0.13, 0.15] \\
Full - dither & 300 & HPSv3 & \tblfocus{-1.41} [-1.78, -1.07] \\
Full - dither & 300 & Objects & 0.03 [-0.00, 0.06] \\
\midrule
FlowEdit: dither - means & 100 & $S_2$ & \tblfocus{0.27} [0.18, 0.37] \\
FlowEdit: dither - means & 100 & $\Delta E$ & -0.01 [-0.13, 0.11] \\
FlowEdit: dither - means & 100 & Edges & \tblfocus{-0.23} [-0.26, -0.21] \\
FlowEdit: dither - means & 100 & HPSv3 & \tblfocus{4.22} [3.42, 5.02] \\
FlowEdit: dither - means & 100 & Objects & \tblfocus{-0.07} [-0.12, -0.03] \\
\midrule
RF-Inversion: means - dither & 100 & $S_2$ & 0.04 [-0.03, 0.11] \\
RF-Inversion: means - dither & 100 & $\Delta E$ & \tblfocus{-0.35} [-0.44, -0.27] \\
RF-Inversion: means - dither & 100 & Edges & \tblfocus{0.08} [0.06, 0.10] \\
RF-Inversion: means - dither & 100 & HPSv3 & \tblfocus{-0.67} [-1.21, -0.20] \\
RF-Inversion: means - dither & 100 & Objects & 0.01 [-0.04, 0.06] \\
\bottomrule
\end{tabularx}
\end{table*}

%% file: appendxi/theory.tex
\section{Theoretical Derivations}\label{app:theory}
\subsection{Soft Region Operators}\label{app:operators}
The implemented regional operation averages with pooled masks and spreads with the same memberships. With the nonempty-region definitions of \cref{eq:operators},
\begin{equation}
 B=AS=W^{-1}MM^\top,\qquad P=M^\top W^{-1}M.
\end{equation}
Hard, disjoint latent masks give $B=I$ and an orthogonal $P$. Pooling across region boundaries generally changes both properties.

\begin{proposition}\label{prop:geometry}
Let $A^\dagger$ be the Moore--Penrose pseudoinverse, $\Pi=A^\dagger A$, and $Q=I-\Pi$. Then $\Pi$ projects orthogonally onto $\range(A^\top)$, $AQ=0$, and $QS=0$. For any offset $\Delta z$, put $\delta_\Delta=A\Delta z$. The starts in \cref{eq:factorial} satisfy
\begin{equation}
\begin{aligned}
 Az(a,b;\Delta z)&=Az_f+\big[aB+b(I-B)\big]\delta_\Delta,\\
 Qz(a,b;\Delta z)&=Qz_f+bQ\Delta z.
\end{aligned}
\end{equation}
If $A$ has full row rank, $A^\dagger=SB^{-1}$.
\end{proposition}
\begin{proof}
The pseudoinverse identity gives that $A^\dagger A$ is the orthogonal row-space projector and $AQ=0$. Since $A^\top=M^\top W^{-1}$, its range equals that of $S=M^\top$, giving $QS=0$ and $QP=0$. Applying $A$ and $Q$ to the factorial construction proves both formulas. If $A$ has full row rank, $MM^\top$ is invertible and
\begin{equation*}
\begin{aligned}
 A^\dagger&=A^\top(AA^\top)^{-1}\\
 &=M^\top(MM^\top)^{-1}W=SB^{-1}.
\end{aligned}
\end{equation*}
\end{proof}

For dependent rows, $AA^\dagger$ is the identity on the attainable mean space $\range(A)$. An exact exchange would use $\Pi\Delta z$ in place of $P\Delta z$. The experimental mean field preserves $Qz$, while the complementary field can change the region means through $(I-B)\delta_\Delta$. These statements apply to both $\Delta z_p=z_p-z_f$ and $\Delta z_d=z_d-z_f$.

For symmetric positive-semidefinite $G$, define $\norm{U}_G^2=\tr(U^\top GU)$. For any prediction error $e_0$, $\Pi=A^\top(AA^\top)^\dagger A$ gives the general regional norm identity
\begin{equation}
\begin{aligned}
 &\norm{\Pi e_0}_F^2\\
 &\quad=\tr\!\left[(Ae_0)^\top(AA^\top)^\dagger(Ae_0)\right].
\end{aligned}
\end{equation}
If $A$ has full row rank, $(AA^\top)^{-1}=W(MM^\top)^{-1}W=WB^{-1}$ is symmetric positive definite, and $\norm{\Pi e_0}_F^2=\norm{Ae_0}_{WB^{-1}}^2$. The matrix $B$ is self-adjoint in the $W$-weighted inner product. Channel projections commute with spatial projections because they act on different axes. With full-row-rank $A$, writing $E=Ae_0$ and $G=WB^{-1}$ therefore yields $\norm{E}_G^2=\norm{Euu^\top}_G^2+\norm{E(I_C-uu^\top)}_G^2$. The reported regional RMS probes use area weights, as defined in \cref{app:probe-metrics}. Orthogonality $\Pi Q=0$ and $e_0=-\sigma e_v$ give the velocity-loss split in \cref{eq:loss-split}.

The nominal dose and actual initial photo-relative position also differ:
\begin{equation}
 \pi_r(z_f+\alpha S\delta)=
 \alpha\frac{\ip{(B\delta)_r}{\delta_r}}{\norm{\delta_r}^2}.
\end{equation}
The reported $\kappa$ is the fitted response to $\alpha$. Its numerical proximity to one is an empirical property of this field intervention and the tested trajectories.

\subsection{Nonlinear Frequency-to-Mean Response}\label{app:encoding}
For a fixed mask, region, and channel, let $f(y)=[A\enc(y)]_{rj}$. Assume $\E\xi=0$, $\E[\xi\xi^\top]=\Sigma$, and $\E\norm\xi^3<\infty$. For $\eta\geq0$, suppose $f$ is three times differentiable along every segment $d+t\eta\xi$, $0\leq t\leq1$, in the perturbation support, with third-derivative operator norm uniformly bounded by $K$. Taylor's theorem gives the remainder in \cref{eq:curvature} as
\begin{equation}
 |R(d,\eta\xi)|\leq\frac{K}{6}\eta^3\norm{\xi}^3.
\end{equation}
Taking expectations eliminates the linear term, and $\E[\xi^\top H\xi]=\tr(H\Sigma)$ gives the quadratic term. Clipping and quantization in practical dither change its input distribution; the local smooth expansion describes the ideal perturbation before those operations. The sign of the leading shift is determined by $\tr(H_f(d)\Sigma)$.

High-pass filtering followed by squaring and averaging isolates the conversion of fine variation into a mean shift, related to the role of nonlinearity and averaging in scattering representations~\cite{bruna2013scattering}. A vector-valued example on a periodic scalar region uses $H\mathbf1=0$, a baseline $b(c)\in\R^C$, and defines
\begin{equation}
 F_{\rm toy}(c\mathbf1+h)=b(c)+u\frac1N\sum_i(Hh)_i^2.
\end{equation}
For a discrete Fourier sinusoid $h_i=\eta\cos(\omega i)$ at a nonzero, non-Nyquist frequency, the displacement equals $u\eta^2|\widehat H(\omega)|^2/2$. Fine-frequency energy thus appears in an averaged statistic after a nonlinearity. A saturating function of this energy yields a saturating mean response.

For an affine encoder, a random perturbation with zero expectation produces zero expected output displacement. A linear strided or boundary-sensitive map can convert a deterministic zero-regional-mean image into a nonzero regional output. The expectation condition in the expansion is therefore essential. Shared-direction alignment with photo offsets is measured independently of this mathematical example.

\subsection{Flow Matching and the Clean Prediction}\label{app:posterior}
The ideal squared-loss velocity regressor is a conditional expectation. Conditional on text $c$, let $Z_0$ and $\varepsilon\sim\mathcal N(0,I)$ be independent. Set $Z_\sigma=(1-\sigma)Z_0+\sigma\varepsilon$ and $V=\varepsilon-Z_0$. Since $Z_0=Z_\sigma-\sigma V$, conditioning gives the velocity and clean prediction
\begin{align}
 v^*(z,\sigma,c)&=\E[V\mid Z_\sigma=z,c],\\
 \widehat z_0^*(z,\sigma,c)&:=z-\sigma v^*(z,\sigma,c)\notag\\
 &=\E[Z_0\mid Z_\sigma=z,c].\label{eq:clean-prediction}
\end{align}
For $0<\sigma<1$ and marginal density $p_\sigma$, differentiation under the Gaussian convolution yields
\begin{equation}
 \E[Z_0\mid Z_\sigma=z,c]
 =\frac{z+\sigma^2\nabla_z\log p_\sigma(z\mid c)}{1-\sigma}.
\end{equation}
This is the corresponding scaled form of Tweedie's formula~\cite{efron2011tweedie}. It describes one clean prediction. A complete deterministic trajectory integrates a changing vector field and is a separate map.

\subsection{Noise Averaging and Gaussian Retention}\label{app:retention-model}
The exact covariance of region-averaged independent noise is $AA^\top$. In particular,
\begin{equation}
\begin{aligned}
 \Var[(A\varepsilon)_r]&=\frac{\sum_iM_{ri}^2}{a_r^2},\\
 \Cov[(A\varepsilon)_r,(A\varepsilon)_s]&=
 \frac{\sum_iM_{ri}M_{si}}{a_ra_s}.
\end{aligned}
\end{equation}
For two references separated by $\Delta q_r$ in a unit channel coordinate, their separation relative to the forward-noise standard deviation is
\begin{equation}
 d'_r=\frac{(1-s)|\Delta q_r|}{s\sqrt{\nu_r}}.
\end{equation}
This measures detectability of a start-coordinate difference. The generator's endpoint response requires a model of the field using that observation.

\begin{proposition}\label{prop:flow}
Assume an autonomous scalar path with $q_0\sim\mathcal N(\mu,\tau^2)$, independent $\xi\sim\mathcal N(0,\nu)$, $\tau,\nu>0$, and $q_\sigma=(1-\sigma)q_0+\sigma\xi$. Let $V_\sigma=(1-\sigma)^2\tau^2+\sigma^2\nu$. For $0<s<1$, at the reference start $q_s^{\rm ref}=(1-s)q_{\rm ref}+s\xi$, the posterior-mean and ideal marginal-flow endpoint sensitivities are
\begin{equation}
\begin{aligned}
 \kappa_{\rm post}&=\frac{(1-s)^2\tau^2}{V_s},\\
 \kappa_{\rm flow}&=\frac{(1-s)\tau}{\sqrt{V_s}}
                 =\sqrt{\kappa_{\rm post}}.
\end{aligned}
\end{equation}
Both derivatives hold the start noise fixed.
\end{proposition}
\begin{proof}
Gaussian conditioning gives
\begin{equation}
 \E[q_0\mid q_s]
 =\mu+\frac{(1-s)\tau^2}{V_s}\big(q_s-(1-s)\mu\big).
\end{equation}
Substituting $q_s^{\rm ref}$ and differentiating proves the first claim. The conditional-mean velocity of the Gaussian path is
\begin{equation}
 v(q,\sigma)=-\mu+\frac{V_\sigma'}{2V_\sigma}
 \big(q-(1-\sigma)\mu\big).
\end{equation}
The standardized coordinate $r_\sigma=(q_\sigma-(1-\sigma)\mu)/\sqrt{V_\sigma}$ has derivative zero along this ODE. Because $V_0=\tau^2$, integration to zero gives
\begin{equation}
\begin{aligned}
 q_{\rm out}
 &=\mu+\frac{\tau}{\sqrt{V_s}}\big(q_s^{\rm ref}-(1-s)\mu\big)\\
 &=\mu+\kappa_{\rm flow}(q_{\rm ref}-\mu)
 +\frac{s\tau}{\sqrt{V_s}}\xi.
\end{aligned}
\end{equation}
Differentiation proves the endpoint formula.
\end{proof}

Taking $\nu=\nu_r$ and $\mathrm{SNR}=(1-s)^2\tau^2/(s^2\nu_r)$ yields \cref{eq:gaussian-retention}. The single-prediction sensitivity is $\kappa_{\rm post}=\mathrm{SNR}/(1+\mathrm{SNR})$, so $\kappa_{\rm flow}=\sqrt{\kappa_{\rm post}}>\kappa_{\rm post}$ under the proposition's assumptions. Both approach one at high SNR: the full Gaussian flow retains a regional coordinate more strongly than one conditional-mean prediction.

The prior variance $\tau^2$ describes conditional distributional variation. Both $\tau^2$ and effective region size determine SNR: at $n_{\rm eff}=21$ and $s=0.82$, $\mathrm{SNR}\approx1.012\tau^2$. Here $\tau^2$ remains an analytical parameter; estimating it for the tested backbones is an open measurement problem. The result assumes an autonomous Gaussian coordinate and an ideal field. For coupled, guided, nonlinear generators, \cref{tab:retention} reports measured endpoint sensitivity and the explanatory power of region area.

\subsection{Appearance Evidence in a Mixture Model}\label{app:appearance-mixture}
A conditional mixture illustrates how an appearance coordinate can compete with a text prior. Let a latent appearance label $K\in\{\mathrm F,\mathrm T\}$ have probabilities $p_k(c)>0$, and suppose $q_0\mid K=k,c\sim\mathcal N(\mu_k,\tau^2)$, with $\mu_{\mathrm T}>\mu_{\mathrm F}$. Under the same independent Gaussian forward noise, put $\Delta=\mu_{\mathrm T}-\mu_{\mathrm F}$. The component log odds are
\begin{equation}
\begin{aligned}
 \Lambda(q_s,c)
 &:=\log\frac{\Pr(K=\mathrm T\mid q_s,c)}{\Pr(K=\mathrm F\mid q_s,c)},\\
 &=\log\frac{p_{\mathrm T}(c)}{p_{\mathrm F}(c)}
 +\frac{(1-s)\Delta}{V_s}(q_s-\overline\mu_s).
\end{aligned}
\end{equation}
Here $\overline\mu_s=\tfrac12(1-s)(\mu_{\mathrm T}+\mu_{\mathrm F})$ is the midpoint of the two noised component means.
Changing the reference coordinate by $h$ changes these log odds by $(1-s)^2\Delta h/V_s$ at fixed noise. Thus a reference statistic can favor a flat appearance even when the text prior favors texture. Posterior component evidence is distinct from the deterministic flow endpoint.

\subsection{Mean Guidance as Gradient Descent}\label{app:guidance-derivation}
Mean guidance minimizes an area-weighted least-squares objective. For a fixed target $m^\star$, the identity $A^\top W=S$ gives
\begin{equation}\label{eq:guidance-objective}
\begin{aligned}
 J(z)&=\frac12\norm{Az-m^\star}_W^2,\\
 \nabla J(z)&=S(Az-m^\star).
\end{aligned}
\end{equation}
At $z=\widehat z_0$, the update $z'=z-\gamma\nabla J(z)$ is exactly \cref{eq:guidance}. It preserves $Qz$ because $QS=0$. Its mean error obeys $e'=(I-\gamma B)e$, with $e=Az-m^\star$.

The Hessian is $P=M^\top W^{-1}M$. For a scalar vector $v$, weighted Cauchy--Schwarz gives
\begin{equation}
\begin{aligned}
 v^\top Pv
 &=\sum_r\frac{(\sum_iM_{ri}v_i)^2}{a_r}\\
 &\leq\sum_{r,i}M_{ri}v_i^2=\norm v^2.
\end{aligned}
\end{equation}
Thus $P$ has eigenvalues in $[0,1]$. For an attainable target, choose $z^\star$ with $Az^\star=m^\star$. Each positive-eigenvalue component of $z-z^\star$ is multiplied by $1-\gamma\lambda$, whose absolute value is below one for $0<\gamma<2$. Isolated steps therefore converge to the target means while preserving $Qz$. Our target is attainable since $m^\star=A[z_f+\alpha(z_d-z_f)]$. This proof concerns a fixed objective; interleaved nonlinear sampling is a separate dynamical system.